\documentclass[10pt,paper,twoside,web]{IEEEtran}

\usepackage{generic}
\usepackage{color}
\usepackage{cite}
\usepackage{amsmath,amssymb,amsfonts}
\usepackage{algorithmic}
\usepackage{graphicx}
\usepackage{textcomp}
\usepackage{subfigure}
\usepackage{ulem}
\usepackage{caption}
\usepackage{float} 
\usepackage{tabularx}

\newcommand{\eproof}{\hfill\rule{2mm}{2mm}}

\newcommand{\bstate}{\medskip\begin{state} }
	\newcommand{\estate}{ \hfill  \rule{1mm}{2mm}\medskip\end{state}}

\newcommand{\bass}{\medskip\begin{ass} }
	\newcommand{\eass}{ \hfill  \rule{1mm}{2mm}\medskip\end{ass}}

\newcommand{\brem}{\medskip \begin{remark}  }
	\newcommand{\erem}{\hfill \rule{1mm}{2mm}\medskip
\end{remark} }
\newcommand{\bthm}{\medskip\begin{theorem}  }
	\newcommand{\ethm}{ \hfill  \rule{1mm}{2mm} \medskip
\end{theorem} }
\newcommand{\blem}{\medskip\begin{lemma}  }
	\newcommand{\elem}{ \hfill \rule{1mm}{2mm}\medskip
\end{lemma} }
\newcommand{\bcorollary}{\medskip\begin{corollary}  }
	\newcommand{\ecorollary}{  \hfill \rule{1mm}{2mm}\medskip
\end{corollary} }
\newcommand{\bdefn}{\medskip\begin{definition}}
	\newcommand{\edefn}{  \hfill \rule{1mm}{2mm}\medskip
\end{definition} }
\newcommand{\bproposition}{\medskip\begin{proposition} }
	\newcommand{\eproposition}{\hfill \rule{1mm}{2mm}\medskip
\end{proposition} }
\newcommand{\bexample}{\medskip\begin{example} \rm}
	\newcommand{\eexample}{ \hfill \rule{1mm}{2mm}\medskip
\end{example} }

\newcommand{\bcon}{\medskip\begin{condition} \rm}
	\newcommand{\econ}{ \hfill \rule{1mm}{2mm}\medskip
\end{condition} }

\renewcommand{\t}{^{\top}}
\newcommand{\proofnow}{\noindent{\bf Proof: }}
\newcommand{\prooflater}[1]{\noindent{\bf Proof of #1: }}

\newtheorem{theorem}{\bf Theorem}[section]
\newtheorem{ass}{\bf Assumption}[section]
\newtheorem{lemma}{\bf Lemma}[section]
\newtheorem{definition}{\bf Definition}[section]
\newtheorem{remark}{\bf Remark}[section]
\newtheorem{corollary}{\bf Corollary}[section]
\newtheorem{proposition}{\bf Proposition}[section]
\newtheorem{example}{\bf Example}[section]
\newtheorem{condition}{\bf Condition}[section]
\newtheorem{state}{\bf Assumption}[section]

\begin{document}
\title{A Three-Level Whole-Body Disturbance Rejection Control Framework for Dynamic Motions in Legged Robots
\author{Bolin Li, Gewei Zuo, Zhixiang Wang, Xiaotian Ke, Lijun Zhu, and Han Ding}
\thanks{This work was supported in part by the National Natural Science Foundation of China under Grant U25A6013, Grant 62173155 and Grant 52188102. (Corresponding author: Lijun Zhu.)}
\thanks{Bolin Li, Gewei Zuo, Zhixiang Wang, and Xiaotian Ke are with the School of Artificial Intelligence and Automation, Huazhong University of Science and Technology, Wuhan 430074,	China
(e-maisl: bolin\_li@hust.edu.cn; gwzuo@hust.edu; zhixiangwang@hust.edu.cn; m202373636@hust.edu.cn).}
\thanks{Lijun Zhu is with the School of Artificial Intelligence and Automation,
	Huazhong University of Science and Technology, Wuhan 430074, China, and
	also with the State Key Laboratory of Intelligent Manufacturing Equipment and
	Technology, Huazhong University of Science and Technology, Wuhan 430074,
	China (e-mail: ljzhu@hust.edu.cn).}
\thanks{Han Ding is with the State Key Laboratory of Intelligent Manufacturing
	Equipment and Technology, Huazhong University of Science and Technology,
	Wuhan 430074, China, and also with the School of Mechanical Science and
	Engineer, and the State Key Laboratory of Intelligent Manufacturing Equipment
	and Technology, Huazhong University of Science and Technology, Wuhan
	430074, China (e-mail: dinghan@hust.edu.cn).}}


\maketitle
\begin{abstract}
This paper presents a control framework designed to enhance the stability and robustness of legged robots in the presence of uncertainties, including model uncertainties, external disturbances, and faults. The framework enables the full-state feedback estimator to estimate and compensate for uncertainties in the whole-body dynamics of the legged robots. First, we propose a novel moving horizon extended state observer (MH-ESO) to estimate uncertainties and mitigate noise in legged systems, which can be integrated into the framework for disturbance compensation. Second, we introduce a three-level whole-body disturbance rejection control framework (T-WB-DRC). Unlike the previous two-level approach, this three-level framework considers both the plan based on whole-body dynamics without uncertainties and the plan based on dynamics with uncertainties, significantly improving payload transportation, external disturbance rejection, and fault tolerance. Third, simulations of both humanoid and quadruped robots in the Gazebo simulator demonstrate the effectiveness and versatility of T-WB-DRC. Finally, extensive experimental trials on a quadruped robot validate the robustness and stability of the system when using T-WB-DRC under various disturbance conditions.      
\end{abstract}

\def\abstractname{Note to Practitioners}
\begin{abstract}
This paper presents a practical control framework to significantly improve the robustness of legged robots against real-world uncertainties like unknown payloads, external pushes, and actuator faults. Its core is a novel three-level whole-body controller (T-WB-DRC) that uses a moving horizon estimator (MH-ESO) to accurately identify and compensate for disturbances in real-time. This dual-planning approach, which considers both ideal and disturbance-injected dynamics, outperforms previous methods. The framework's effectiveness in enhancing stability under disturbances has been successfully validated through extensive simulations and physical experiments on a quadruped robot.
\end{abstract}


\section{Introduction}
\IEEEPARstart{L}{egged} robots have attracted considerable attention in recent years due to their capability to traverse complex and uneven terrains \cite{wisth2022vilens}, providing significant benefits across various applications, including search and rescue, exploration, and assistive robotics. However, a major challenge in legged robot locomotion is maintaining stability and robustness in the presence of external disturbances and model uncertainties, such as uneven ground \cite{meng2023online,10766928,8672785}, unforeseen external forces, or heavy payloads \cite{sun2021online}. These disturbances can severely impact the robot’s state, causing deviations from the desired trajectory and ultimately undermining both performance and stability.

To improve the stability and robustness of legged robots, various disturbance rejection methods have been developed and implemented. One approach involves designing sophisticated controllers, such as limit cycle walkers \cite{10335933} and high-performance optimal control \cite{sombolestan2024adaptive, 10384795}.  In \cite{sombolestan2024adaptive}, a novel methodology for integrating adaptive control into a force-based control system was proposed for agile-legged robots. In \cite{xu2023robust}, a min-max optimization model based on robust optimization was formulated, and a robust min-max model predictive controller was proposed to address quadruped locomotion control under uncertainties, specifically uncertain friction constraints and model dynamics. In \cite{grandia2023perceptive}, a two-level framework consisting of high-level and low-level control was proposed for robust payload transportation by quadrupedal robots. These methods effectively tackle model uncertainties in quadruped robots during payload transportation, especially those associated with the single rigid body model. Another approach leverages disturbance rejection observers to further strengthen the robot’s robustness. In \cite{zhu2023proprioceptive}, a control framework for legged robots that enables self-perception and resilience to external disturbances was introduced, integrating a novel proprioceptive-based disturbance estimator that offers significant advantages in filtering foot-ground interaction noise and preventing the accumulation of estimation errors.  In \cite{bledt2018contact}, a discrete-time generalized-momentum disturbance observer is used to enhance proprioceptive force control estimates, which are then combined with contact priors using Kalman Filtering for greater robustness. The generalized-momentum disturbance observer effectively estimates external disturbances in legged systems, but it encounters challenges when dealing with model uncertainties, particularly in robots with significant friction and complex dynamic interactions \cite{11079307}. Similar to model uncertainties and external disturbances, fault conditions also reduce the robustness of legged robots due to susceptibility to various factors. Previous work on fault-tolerant locomotion in legged robots can be broadly classified into trajectory planning \cite{chen2024fault} and learning-based approaches \cite{luo2023ft}. However, few instances have been found where estimators are used to address fault tolerance in legged robots, due to the limited application of full-state observers, despite the widespread use of estimators to improve fault tolerance \cite{huang2020disturbance,ren2024fixed}.

The work in \cite{11079307} proposed a two-level whole-body disturbance rejection control framework (WB-DRC) that integrates an extended state observer (ESO) with adaptive control and a moving horizon filter (MAF) to estimate uncertainties. This framework provides an opportunity to enhance fault tolerance capability in legged robots by utilizing a full-state estimator. The framework compensates for these uncertainties in the low-level control by solving a quadratic programming (QP) optimization problem, simultaneously addressing fault tolerance, model uncertainties, and external disturbances in legged systems. 
However, due to the sensitivity of the ESO to noise \cite{sun2021composite,prasov2012nonlinear}, the performance of the adaptive ESO in estimating uncertainties within the WB-DRC is compromised as noise increases, particularly with the adoption of a larger moving horizon window in the MAF or a smaller ESO bandwidth. If uncertainty estimation performance is not compromised, the QP problem may still become infeasible due to the influence of noise.

In this paper, we propose a three-level framework that incorporates a full-state feedback estimator and a single processing module at the mid-level to estimate uncertainties in legged systems, addressing model uncertainties, external disturbances, and faults. This framework can include not only the ESO but also other full-state estimators, such as predefined-time observers and model reference adaptive estimators to estimate the uncertainties in legged systems. Furthermore, to mitigate the impact of noise on the legged system, we propose an MH-ESO, which is less sensitive to noise than the ESO, to act as the full-state estimator within the three-level framework. A robust model predictive control, which is less sensitive to noise than QP solving, is adopted to plan the trajectories of dynamics with uncertainties for low-level control. The main contributions and important emphases are the following:

1) A three-level whole-body disturbance rejection control framework is proposed to enhance the robustness of legged robots in the presence of model uncertainties, external disturbances, and faults. Full-state feedback estimators, including ESO and MH-ESO, can be employed within the framework to estimate uncertainties in legged robots, which are hybrid systems. Additionally, commonly used filters for smoothing interaction noise, such as MAF and low-pass filters, can be incorporated into the framework.

2) A moving horizon extended state observer, which is less sensitive to noise than the ESO, is proposed. The moving horizon extended state observer's fast computation is achieved by solving the linear equation.

3) Simulations are conducted on both humanoid and quadruped robots, and experiments are performed on the Unitree A1 quadruped robot to validate the effectiveness of the proposed T-WB-DRC, highlighting its advantages in robustness and stability.

This paper is organized as follows. Section \ref{sec:preliminaries} provides background information on legged robots. Section \ref{sec:High_level_Control} presents the high-level control approach. Section \ref{sec:uncertainties_estimation} introduces the uncertainties estimator. Section \ref{sec:robust_mpc} presents the robust model predictive control. Section \ref{sec:low_level_control} provides the low-level control.  Sections \ref{sec:simulation_comparison} and \ref{sec:experimental_verification} assess the performance of the framework through simulations and experimental verification, respectively. Finally, Section \ref{sec:conclusion} concludes the paper.

\section{Preliminaries}\label{sec:preliminaries}
A legged robot is a free-floating-base system composed of
$6+n$ degrees of freedom 
\begin{gather}
	D(q)\ddot{q} + C(q,\dot{q})\dot{q} + G(q) = S\t \tau + J(q)\t F + d\label{eq:dynamics}
\end{gather}
where $q = [(q^b)\t, (q^j)\t]\t$ with $q^b \in \mathbb{R}^6$, $D(q) : \mathbb{R}^{6+n}\to \mathbb{R}^{(6+n)\times(6+n)}$ is a positive definite inertia matrix, $C(\dot{q}, q) :\mathbb{R}^{6+n} \times \mathbb{R}^{6+n} \to \mathbb{R}^{(6+n)\times(6+n)}$ is the Coriolis matrix, $G(q) : \mathbb{R}^{6+n} \to \mathbb{R}^{6+n}$ is the gravity vector;
$S = [0_{n \times 6}\;\;I_n]$ is the actuated part selection matrix; $\tau \in \mathbb{R}^{n}$ is the actuation torques; $F \in \mathbb{R}^{3 n_c}$ is the ground reaction forces (GRFs) of the stance leg; $J \in \mathbb{R}^{3n_c \times (6+n)}$ is the Jacobians that transposes the GRFs into the acceleration of the
base and the actuated joints; $d \in \mathbb{R}^{n+6}$ is the system uncertainty; $n$ and $n_c$ represent the number of active degrees of freedom and the number of legs on the support surface, respectively.

We denote $h$ as the linear momentum and ${L}$ as the angular momentum around the robot's center of mass (CoM). Let   $c \in \mathbb{R}^{3}$ be the position of the CoM in the inertial frame, and $h = m\dot{c}$ be the linear momentum, where $m$ is the total mass of the robot. The centroidal momentum is linked to the joint configuration and its derivatives through the centroidal mapping
\begin{gather}
\left[ {\begin{array}{*{20}{c}}
		h\\
		L
\end{array}} \right] = A(q)\dot{q}
\end{gather}
with $A(q)$ being the so-called centroidal momentum matrix \cite{orin2013centroidal}. The centroidal dynamics \cite{orin2008centroidal} can be obtained according to the Newton-Euler law:
\begin{align}
\begin{array}{l}
	\dot h = \sum\limits_{i = 1}^{{n_c}} {{f_i}}  - mg\\
	\dot L = \sum\limits_{i = 1}^{{n_c}} {({p_i} - c) \times {f_i}} 
\end{array} \label{eq:centroidal_dynamics}
\end{align}
where $f_i \in \mathbb{R}^3$ represents the contact force of contact $i$, $p_i$ denotes the position of contact $i$, and $g = [0,0,9.81]\t$ is the gravity vector. The centroidal dynamics (\ref{eq:centroidal_dynamics}) is referred to as the nominal centroidal dynamics, where the force $f_i$ is considered only as the ground reaction force, with external forces neglected, and $m$ represents the robot's nominal mass, excluding model uncertainties.

\begin{figure}[!ht]
	\centering
	\includegraphics[scale=0.48]{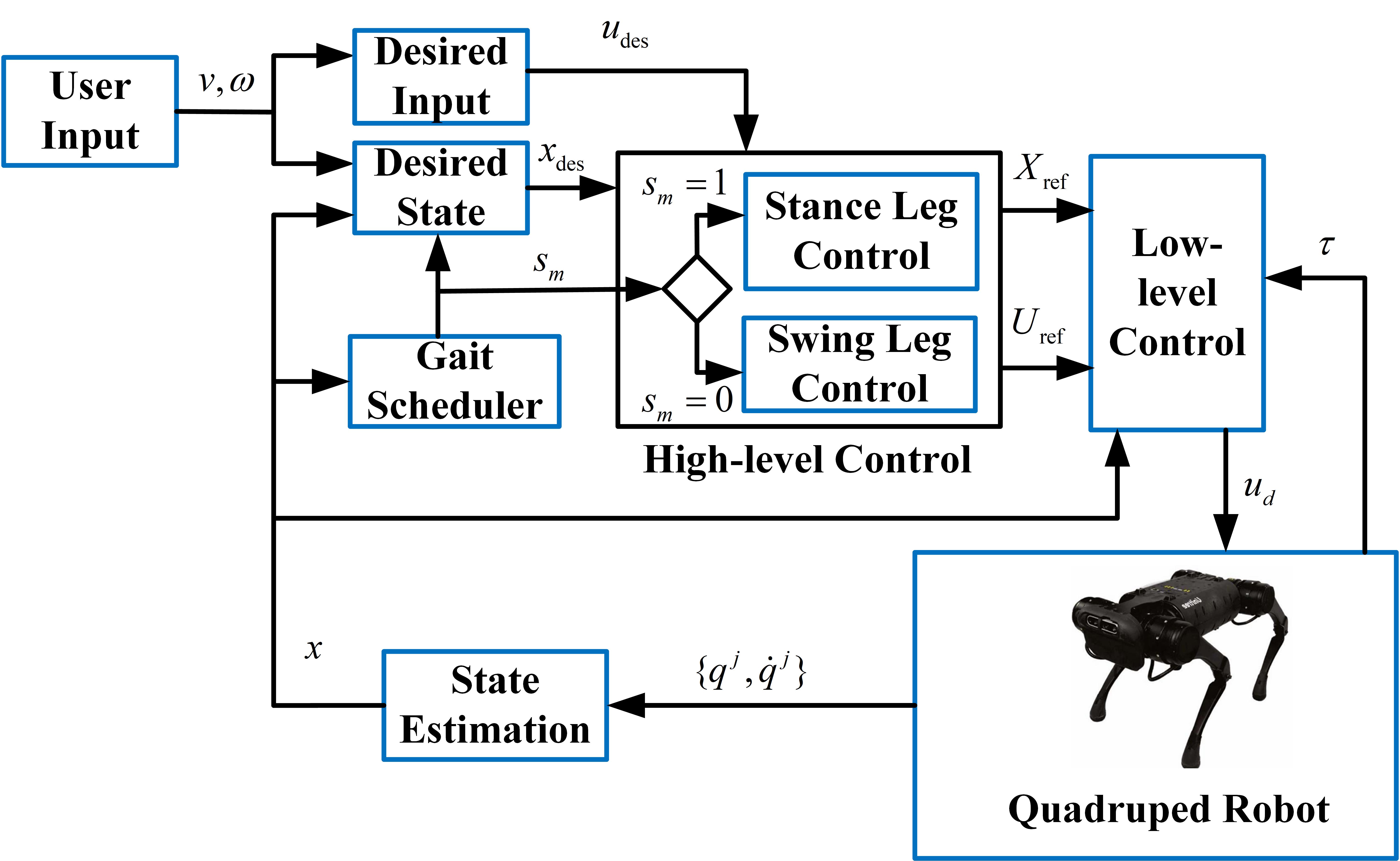}
	\caption{A block diagram of a typical two-level control framework for legged robots.}
	\label{fig_control_structure_1}
\end{figure}

The     legged robot’s control system normally consists of several modules, including a two-level control (high-level control and low-level control), state estimation, and gait scheduler \cite{bledt2018cheetah}, as presented in Fig. \ref{fig_control_structure_1}. A reference trajectory can be generated by the high-level control based on the user input and the state estimation. The gait scheduler determines the timing and sequence of the gait, coordinating the transition between the swing and stance phases of each leg. The high-level control module manages to generate the trajectory for the swing legs and the optimal GRFs for the stance legs, all in accordance with the user commands and gait timing. These components of the control architecture are detailed in \cite{di2018dynamic,focchi2017high,bledt2018cheetah}. 

\begin{figure}[!ht]
	\centering
	\includegraphics[scale=0.46]{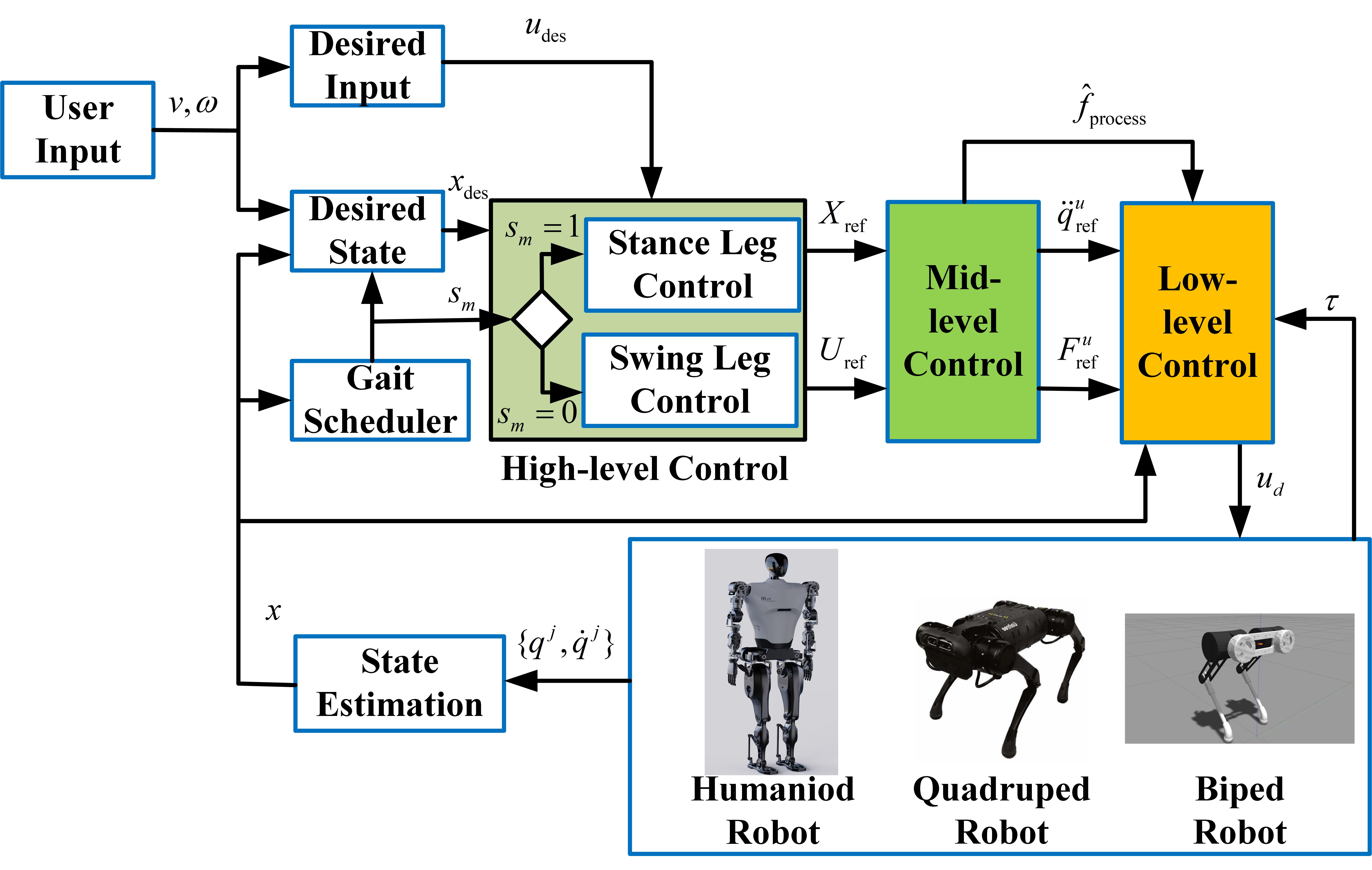}
	\caption{Block diagram of the proposed T-WB-DRC.}
	\label{fig_control_structure}
\end{figure}

\begin{figure}[!ht]
	\centering
	\includegraphics[scale=0.50]{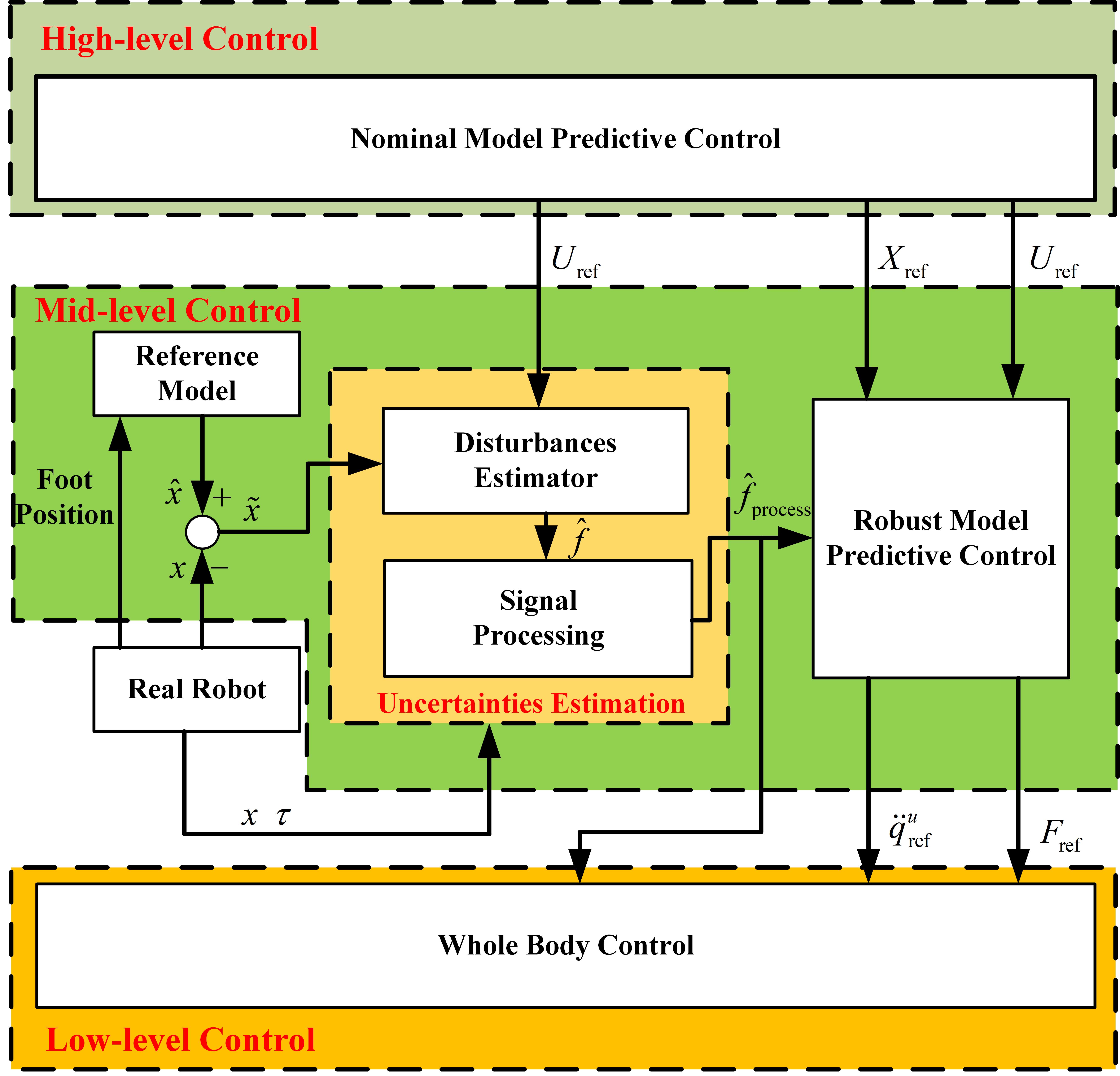}
	\caption{Block diagram of three-level control in T-WB-DRC.}
	\label{fig_mid_level}
\end{figure}

To address the uncertainties and noise in legged robots, we propose a three-level whole-body disturbance rejection control architecture, as illustrated in Fig. \ref{fig_control_structure}. The details of the three-level control framework are further illustrated in Fig. \ref{fig_mid_level}. We employ model predictive control (MPC) in the high level to plan the legged robot's centroidal momentum, joint positions, joint velocities, and GRFs based on nominal centroidal dynamics. The MH-ESO in middle level  estimates disturbances and addresses the noise in the legged system, while the MPC is then used to replan the robot's joint acceleration and GRFs, based on the reference trajectory from high-level control.
Finally, the whole-body control in the low level   compensates for system disturbances.
{
The proposed framework uses an estimator to estimate and compensate for model uncertainties, external disturbances, and faults. However, it treats these challenges the same without considering their individual effects. Internal model uncertainty requires the estimator to distinguish between parameter drift and true state changes, while external disturbances are easier to estimate if their dynamics differ from the system. Measurement noise creates a trade-off between the observer's bandwidth and its ability to suppress noise. Sudden faults put the most pressure on the estimator, requiring fast detection and compensation to keep the system stable.

}

\section{High-level Control} \label{sec:High_level_Control}
The high-level control employs MPC to generate reference trajectories for the GRFs, the robot's joint positions, joint velocities, and the centroidal momentum for the mid-level control.
The same high-level control presented in \cite{11079307} is adopted in this paper, and we revisit it in this section. We use the centroidal dynamics (\ref{eq:centroidal_dynamics}) for the high-level planning. Denote $x_c = [h\t, L\t, q\t]\t$, $u_c = [F\t, (\dot{q}^j)\t]\t$ with $F = [f\t_1,\cdots,f\t_{n_c}]\t \in \mathbb{R}^{3n_c}$. 
The centroidal dynamics (\ref{eq:centroidal_dynamics}) can be rewritten as
\begin{gather}
	\dot{x}_c = f_c(x_c,u_c)\label{eq:centroidal_dynamics_1}
\end{gather}
The continuous dynamics (\ref{eq:centroidal_dynamics_1}) is discretized over intervals of the prediction horizon $[t,t+T]$ for the nonlinear trajectory optimization problem. We consider the total number of steps in the MPC to be $H$, and thus the duration of each discrete interval is $\delta_t = T/(H-1)$. Define $J_c(x)$ as the terminal quadratic cost around the reference state $x_{\rm{des}}$
\begin{gather}
J_c(x) = (x-x_{\rm{des}}(t+T))\t Q_t (x-x_{\rm{des}}(t+T))
\end{gather}
and $\varphi_{c,i}(x,u)$ as the quadratic cost around the reference state $x_{\rm{des}}(t)$ and input $u_{\rm{des}}(t)${
\begin{align}
	\varphi_{c,i}(x,u) =& (x-x_{\rm{des}}(t + i\delta_t))\t Q_s (x-x_{\rm{des}}(t+i\delta_t)) \nonumber\\&+ (u - u_{\rm{des}}(t+i\delta_t))\t R_s (u - u_{\rm{des}}(t+i\delta_t))\nonumber	
\end{align}
for $i = 1,\cdots,H - 1$.}
Let $x_{i,k}$ and $u_{i,k}$ denote the sequences of state and input variables in the MPC at the $k$-th solve, respectively. The nonlinear MPC problem can be formulated by defining and evaluating a cost function and constraints as \cite{11079307,grandia2023perceptive}
	\begin{align}
		\text{min}_{X, U}\;\;\;\;\;\;\;&J_{c}(x_{H,k}) + \sum ^{H-1}_{i=0} \varphi_{c,i}(x_{i,k}, u_{i,k}) + l_i(x_{i,k},u_{i,k})\nonumber\\
		\mbox{s.t.} \;\;\;\;\;\;\; &x_{0,k} - {x}_{m,k} = 0,\nonumber\\
		\;\;\;\;\; &x_{i+1,k} - x_{i,k} - f_c(x_{i,k},u_{i,k})\delta t =0, \nonumber\\
		\;\;\;\;\; &g_i(x_{i,k},u_{i,k}) = 0,\;\; i = 0,\cdots,H-1. \label{eq:MPC_problem}
	\end{align}
where ${x}_{m,k}$ is the current measured state,  $X = [x_{0,k}\t, \cdots, x_{H,k}\t]\t$ and $U = [u_{0,k}\t,\cdots,u_{H-1,k}\t]\t$ are the sequences of state and input variables respectively. The penalized cost function $l_i$ is derived from inequality constraints, which ensure that the swing leg remains above the ground and that the contact force lies within the friction cone. The constraint function $g_i$ represents the equality requirements, ensuring that the swing leg's end does not experience a contact force and that the end of the stance leg has zero velocity. Note that the legged robot is a hybrid system, when managed by the gait scheduler,  $f_i$ in the GRFs $F$ is either a zero vector or a nonzero vector.

The centroidal dynamics (\ref{eq:centroidal_dynamics_1}) represents the nominal centroidal dynamics. We use the nominal dynamics in high-level control, as we prefer to generate the reference trajectory under the assumption that the system does not encounter disturbances. The solution to the   MPC problem (\ref{eq:MPC_problem}) is denoted as 
$X=X_{\rm{ref}} := [h_{\rm{ref}}\t, {L}_{\rm{ref}}\t, q_{\rm{ref}}\t]\t$ and  $U=U_{\rm{ref}} := [F_{\rm{ref}}\t, (\dot{q}^{j}_{\rm{ref}})\t]\t$, which gives the reference signals at the discrete time instances $[t,t+\delta_t,\cdots, t+T]\t$. 
We can linearly interpolate the discrete  reference signals  to a continuous reference trajectory  for the interval $[t,t+T]$.

\section{Uncertainties Estimation}\label{sec:uncertainties_estimation}
In this section, we introduce two variants of disturbance estimators, namely ESO and MH-ESO,  which are designed based on ESO, to estimate uncertainties in the legged system. The disturbance estimators are  discussed in Section \ref{sec:disturbance_estimator}. Both disturbance estimators can be adopted in our proposed T-WB-DRC framework. We also present the signal processing module in T-WB-DRC in Section \ref{sec:signal_processing}.

\subsection{Disturbance Estimator} \label{sec:disturbance_estimator}
{
The influence of measurement noise on disturbance estimators is always present in legged robots. For almost all disturbance estimators in legged robots, the measurement noise is not directly considered in the estimator design; instead, it is handled in the signal processing module using an MAF filter or low-pass filter. In this section, we will start by designing the ESO without considering the influence of measurement noise and then integrate it into the moving horizon estimation (MHE) process to effectively manage the noise in the estimator design. 
}

Let $x_1 = q$ and $x_2 = \dot{q}$. The dynamics model (\ref{eq:dynamics}) can be written in the state-space form 
\begin{gather}
\left[ {\begin{array}{*{20}{c}}
		{{{\dot x}_1}}\\
		{{{\dot x}_2}}
\end{array}} \right] = \left[ {\begin{array}{*{20}{c}}
		{{x_2}}\\
		{{f_e}({x_1},{x_2})}
\end{array}} \right] + \left[ {\begin{array}{*{20}{c}}
		0\\
		{{g_e}({x_1})}
\end{array}} \right]u + \left[ {\begin{array}{*{20}{c}}
		0\\
		{{d}_e}
\end{array}} \right]\label{eq:wholeBodyDynamics}
\end{gather}
where
\begin{align}
&{d}_e := D^{-1}(x_1)d\nonumber\\
&{f_e}({x_1},{x_2}) :=  - {D^{ - 1}}({x_1})(C({x_2},{x_1})x_2 + G({x_1}))\nonumber\\
&{g_e}({x_1}) := {D^{ - 1}}({x_1})B\nonumber
\end{align}
with $B = [S\t, J(x_1)\t]$ and $u = [\tau\t, F\t]\t$.

\subsubsection{ESO design}
We extend the state with $x_3 \in \mathbb{R}^{6 + n}$. Let $x_3 = d_e$  and the time-varying signal $h_x$ denote the rate of change of uncertainty $d_e$, i.e., $\dot{x}_3=h_x(t)$. $x_3$ and $h_x(t)$ are assumed to be unknown but bounded vector functions. The system (\ref{eq:wholeBodyDynamics}) can be expressed in the  state-space form as
\begin{gather}
\left\{ \begin{array}{l}
	{{\dot x}_1} = {x_2}\\
	{{\dot x}_2} = {f_e}({x_1},{x_2}) + {g_e}({x_1})u + {x_3}\\
	{{\dot x}_3} = h_x(t)
\end{array} \right.\label{eq:state_space_form}
\end{gather}

Let $\hat{x}_1, \hat{x}_2$, and $\hat{x}_3$ be the estimates of $x_1$, $x_2$, and $x_3$, respectively. Referring to \cite{gao2003scaling}, the linear ESO can be constructed as follows
\begin{gather}
\left\{ \begin{array}{l}
	{{\dot {\hat x}}_1} = {{\hat x}_2} + 3{\omega _0}({x_1} - {{\hat x}_1})\\
	{{\dot{ \hat x}}_2} = {f_e}({x_1},{x_2}) + {g_e}({x_1})\hat u + {{\hat x}_3} + 3\omega _0^2({x_1} - {{\hat x}_1})\\
	{{\dot {\hat x}}_3} = \omega _0^3({x_1} - {{\hat x}_1})
\end{array} \right.\label{eq:ESO_1}
\end{gather}
where $\hat{u} = [\tau\t, F_{\rm{ref}}\t]\t$ with $F_{\rm{ref}}$ being the partial solution of  the MPC problem (\ref{eq:MPC_problem}) and $\omega_0$ is tuning parameter to be determined. While a large $\omega_0$ is necessary for fast disturbance estimation, it can also amplify noise. We use $F_{\rm{ref}}$  instead of $F$ in (\ref{eq:ESO_1}), because we would like to estimate the discrepancy between $F$ and $F_{\rm{ref}}$  using the ESO.

Next, we analyze the stability of the linear ESO. Let
\begin{gather}
\eta  = [{({x_1} - {\hat x_1})\t},{({x_2} - {\hat x_2})\t}/{\omega _0},{({x_3} - {\hat x_3})^T}/\omega _0^2]\t
\end{gather}
be the state estimation error and $\tilde{u} = u - \hat{u}$. The dynamics of the state estimation errors can be expressed as 
\begin{gather}
\dot \eta  = {\omega _0}A\eta  + \frac{{{C_1}{g_e}({x_1})\tilde u}}{{{\omega _0}}} + \frac{{{C_2}h_x(t)}}{{\omega _0^2}} \label{eq:error_dynamics}
\end{gather}
where $\tilde{u}$ is an unknown but a bounded vector, due to that $F_{\rm{ref}}$  is, $g_e(x_1)$ is a bounded matrix, and 
\begin{align}
	A =& \left[ {\begin{array}{*{20}{c}}
			{ - 3}&1&0\\
			{ - 3}&0&1\\
			{ - 1}&0&0
	\end{array}} \right] \otimes {I_{6 + n}},\nonumber\\ {C_1} =& \left[ {\begin{array}{*{20}{c}}
{\rm{0}}&{\rm{1}}&{\rm{0}}
\end{array}} \right] \t \otimes {I_{6 + n}},\nonumber\\
	{C_2} =& \left[ {\begin{array}{*{20}{c}}
			{\rm{0}}&{\rm{0}}&1
	\end{array}} \right] \t \otimes {I_{6 + n}}.
\end{align}
Note that the matrix $A$ is Hurwitz. The following assumption is considered.

\bass\label{ass:4.1}
The error vector $g_e\tilde{u}$ is bounded, i.e., $||g_e\tilde{u}|| \le \tilde{u}_b$, where $\tilde{u}_b$ is a constant.
\eass

\brem
The matrix $g_e$ given in (\ref{eq:wholeBodyDynamics}) is bounded because the inverse of the inertia matrix $D^{-1}$ and the Jacobians $J$ are both bounded when the legged robots operate within the workspace \cite{10106431}.  If the legged robot operates without uncertainty, and the robot's position $q$ and velocities $\dot{q}$ perfectly track $q_{\rm{ref}}$ and $\dot{q}_{\rm{ref}}$, then $\hat{u}$ will be equal to $u$, i.e., $||\tilde{u}|| = 0$. This implies that Assumption \ref{ass:4.1} is not a strict requirement, especially when the uncertainties are small.
\erem

\bthm\label{thm}
Consider the error dynamics (\ref{eq:error_dynamics}) under Assumption \ref{ass:4.1}. Suppose both the disturbance $x_3$ and its time derivative $h_x$ are bounded  over time, i.e., $||x_3|| \le d_b$ and $\sup_t|| h_x(t)|| \le h_b$, then the estimation error $\eta$ is bounded.\ethm
The proof of Theorem \ref{thm} is provided in Appendix.

Note that $d_e = D^{-1}d$ in (\ref{eq:wholeBodyDynamics}) is estimated by $\hat{x}_3$ in ESO. The estimated $\hat{f}$ for the uncertainties $d$ in (\ref{eq:dynamics}) are given by 
\begin{gather}
\hat{f} = D(x_1)\hat{x}_3.
\end{gather}

\subsubsection{MH-ESO design}
Denote $y_1$ and $y_2$ as the measurements of the position and velocity, respectively. That is,
\begin{align}
	{y_1} =& {x_1} + {\eta _{\rm{pos}}}\nonumber\\
	{y_2} =& {x_2} + {\eta _{\rm{vel}}}
\end{align} 
where $\eta_{\rm{pos}}$ and $\eta_{\rm{vel}}$ represent the position and velocity measurement noise, respectively. 

Since the states $x_1$ and $x_2$ cannot be measured directly, we express the linear ESO in (\ref{eq:ESO_1}) as follows:
\begin{gather}
\left\{ {\begin{array}{*{20}{l}}
		{{{\dot {\hat x}}_1} = {{\hat x}_2} + 3{\omega _0}({y_1} - {{\hat x}_1}) + {w_1}}\\
		{{{\dot {\hat x}}_2} = {f_e}({y_1},{y_2}) + {g_e}({y_1})\hat u  + {{\hat x}_3} + 3\omega _0^2({y_1} - {{\hat x}_1}) + {w_2}}\\
		{{{\dot{\hat x}}_3} = \omega _0^3({y_1} - {{\hat x}_1}) + {w_3}}
\end{array}} \right. \label{eq:ideal_ESO}
\end{gather}
where 
\begin{align}
	{w_1} =&  - 3{\omega _0}{\eta _{\rm{pos}}}\nonumber\\
	{w_2} =& {f_e}({x_1},{x_2}) - {f_e}({y_1},{y_2}) + ({g_e}({x_1}) - {g_e}({y_1})) - 3\omega _0^2{\eta _{\rm{pos}}}\nonumber\\
	{w_3} =&  - \omega _0^3{\eta _{\rm{pos}}}.\nonumber
\end{align}

\bass \label{ass:ass4_2}
$f_e(x_1,x_2): \mathbb{R}^{6+n} \times \mathbb{R}^{6+n} \to \mathbb{R}^{6+n}$
is globally Lipschitz continuous  with respect to $[x_1\t,x_2\t]\t$. For any two constant vectors $a = [a_1\t,a_2\t]\t \in \mathbb{R}^{12+2n}$ and $b = [b_1\t,b_2\t]\t \in \mathbb{R}^{12+2n}$, $f_e$ satisfies
\begin{gather}
	||{f_e}(a_1,a_2) - {f_e}(b_1,b_2)|| \le \gamma_1 ||a - b||.\nonumber
\end{gather}
where $\gamma_1 > 0$ represents the Lipschitz constant.
\eass

\bass\label{ass:ass4_3}
$g_e(x_1): \mathbb{R}^{6+n} \to \mathbb{R}^{6+n}$ is globally Lipschitz continuous  with respect to $x_1$. For any two constant vectors $a_3,b_3 \in \mathbb{R}^{6+n}$, $g_e$ satisfies
\begin{gather}
	||{g_e}(a_3) - {g_e}(b_3)|| \le \gamma_2 ||a_3 - b_3||.\nonumber
\end{gather}
where $\gamma_2 > 0$ represents the Lipschitz constant.
\eass
Under Assumptions \ref{ass:ass4_2} and \ref{ass:ass4_3}, the vector $w_2$ is bounded, given that the noise terms $\eta_{\rm{pos}}$ and $\eta_{\rm{vel}}$ are bounded, and the stability of the ESO in (\ref{eq:ESO_1}) is equivalent to the stability of the ESO in (\ref{eq:ideal_ESO}). The difference between (\ref{eq:ESO_1}) and (\ref{eq:ideal_ESO}) lies in that the feedback signals $y_1$ and $y_2$ in (\ref{eq:ideal_ESO}) can be measured directly, while $w = [w_1\t,w_2\t,w_3\t]\t$ can be obtained optimally using the following MHE.

 Specifically, we employ MHE to estimate the state $\hat{x}$ in (\ref{eq:ideal_ESO}) and deal with the noise $w$. Let 
\begin{gather}
z(k) =[\hat{x}_1\t(k), \hat{x}_2\t(k), \hat{x}_3\t(k)]\t.\nonumber
\end{gather}
The ESO in (\ref{eq:ideal_ESO}) can be discretized using the backward Euler formula, yielding 
\begin{align}
z(k + 1) =& {A_0}z(k) + {u_z}(k) + {w}_t \label{eq:ESOinMHE}
\end{align}
where $A_0 = \bar{A}_0 - L_0C_0$ and  
\begin{align}
u_z(k) =& B_0[f_e(y_1(k), y_2(k)) + g_e(y_1(k)){\hat{u}(k)}] +L_0y_1(k) \nonumber
\end{align}
with $t_s$ being the sampling period and
\begin{gather}
\bar{A}_0 = \left[ {\begin{array}{*{20}{c}}
		1&t_s&0\\
		0&1&t_s\\
		0&0&1
\end{array}} \right] \otimes {I_{6 + n}},\nonumber\\B_0 =\left[ {\begin{array}{*{20}{c}}
0&{{t_s}}&0
\end{array}} \right] \t \otimes {I_{6 + n}},\nonumber\\
C_0 = \left[ {\begin{array}{*{20}{c}}
		1&0&0
\end{array}} \right] \otimes {I_{6 + n}},\;\;\nonumber\\
L_0 = \left[ {\begin{array}{*{20}{c}}
		{3{\omega _0}{t_s}}&{3\omega _0^2{t_s}}&{\omega _0^3{t_s}}
\end{array}} \right] \t \otimes {I_{6 + n}}.\nonumber
\end{gather}
The $u_z(k)$  is considered as the system input for the discrete system (\ref{eq:ESOinMHE}), with $y_1(k)$, $y_2(k)$, and $\hat{u}(k)$ either being directly measured or obtained through state estimation.

Let 
\begin{gather}
y(k) = [{y}_1\t(k), {y}_2\t(k), 0_{6+n}\t]\t\nonumber
\end{gather}
represent the measurement data used for solving the subsequent MHE problem.
At each step $k = N, N+1,\cdots$, the goal of MHE is to estimate state ${z}(k-N|k)$ in (\ref{eq:ESOinMHE}) as $\hat{z}(k-N|k)$, based on the past measurement data $y^{k}_{k-N} = \text{col}(y(k-N),\cdots,y(k))$, the past input $u^{k-1}_{k-N} = \text{col}(u_z(k-N),\cdots,u_z(k-1))$, and the prediction state $\bar{z}(k-N|k)$ for $z(k-N)$, where $N$ is the moving horizon size. Consider the following cost function:
\begin{align}
J(k) =& ||\hat{z}(k - N|k) - \bar z(k - N|k)||_{\Gamma} ^2 + \sum\limits_{i = k - N}^{k - 1} {||\hat{w}(i|k)||^2_{\Lambda}}  \nonumber\\&+ \sum\limits_{i = k - N}^k {||y(i)  - \hat{y}(i|k)||^2_{\Pi}}.\label{eq:cost_function}
\end{align}
Then, at each step $k = N, N+1, \cdots$, the following MHE optimization problem is solved.

\textbf{Problem 1:} Given a triplet $\{\bar{z}(k-N|k), y^{k}_{k-N}, u^{k-1}_{k-N}\}$, find the optimal estimate $\hat{z}(k-N|k)$ and $\hat{w}^{k-1|k}_{k-N|k}$ that
\begin{gather}
\mathop {\min }\limits_{\hat{z}(k - N|k),\hat{w}_{k - N|k}^{k - 1|k}} J(k)
\end{gather}
with the optimal prediction $\bar{z}$ in $J(k)$ is determined as
\begin{align}
\bar z(0|N) =& \bar z(0)\nonumber\\
\bar z(j + 1|k) =& A_0 \hat{z}(j|k - 1) + u_0(j) + \hat{w}(j|k - 1)\nonumber\\
j =& k - N - 1, \cdots ,k - 2,k - 1, \label{eq:barZ}
\end{align}
and the following constraints are satisfied
\begin{align}
\hat{z}(i+1|k) =& A_0{\hat{z}}(i|k) + u_0(i) +  \hat{w}(i|k)\nonumber\\
\hat{y}(i|k) =& C_0{\hat{z}}(i|k). \label{eq:constriants}
\end{align}

\bthm\label{thm:thm4.1}
Given a triplet $\{\bar{z}(k-N|k), y^{k}_{k-N}, u^{k-1}_{k-N}\}$, the solution to Problem 1 is given by
\begin{align}
\hat{z}(k-N|k) =& (\Gamma + M_1)^{-1}[M_2y^{k}_{k-N} + \Gamma \bar{z}(k-N|k) \nonumber \\ & -M_2 H_N u^{k-1}_{k-N}]\label{eq:hatZOptimal}\\
\hat{w}_{k - N|k}^{k - 1|k} =& {M_3^{ - 1}} \Pi {H_N\t}[y_{k - N}^k - {F_N}\hat z(k - N|k) \nonumber\\ &- {H_N}{u}_{k - N}^{k - 1}]\label{eq:solution}
\end{align}
where
$
{M_1} = \Pi {F_N\t}[I - \Pi {H_N}{M_3^{ - 1}}{H_N\t}]{F_N}$, $
{M_2} = \Pi {F_N\t}[I - \Pi {H_N}{M_3^{ - 1}}{H_N\t}]$,
${M_3} = \Lambda + \Pi {H_N\t}{H_N}
$
with
\begin{small}
\begin{gather}
\begin{array}{l}
	{F_N} = \left[ {\begin{array}{*{20}{c}}
			C_0\\
			{C_0A_0}\\
			{C_0{A_0^2}}\\
			\vdots \\
			{C_0{A_0^N}}
	\end{array}} \right],{H_N} = \left[ {\begin{array}{*{20}{c}}
			0&0& \cdots &0\\
			C_0&0& \cdots &0\\
			{C_0A_0}&C_0& \ddots & \vdots \\
			\vdots & \vdots & \ddots &0\\
			{C_0{A_0^{N - 1}}}&{C_0{A_0^{N - 2}}}& \cdots &C_0
	\end{array}} \right].\nonumber\\
\end{array}
\end{gather}
\end{small}\ethm
\proofnow
From (\ref{eq:constriants}), the optimal observation vectors $\hat{y}^{k}_{k-N} = \text{col}(\hat{y}(k-N), \cdots, \hat{y}(k))$ can be written as follows:
\begin{gather}
	\hat{y}^k_{k-N} = F_N \hat{z}(k-N) + H_N {u}^{k-1}_{k-N} + H_N \hat{w}^{k-1}_{k-N}.\label{eq:haty}
\end{gather}
Substituting (\ref{eq:haty}) into (\ref{eq:cost_function}) gives
\begin{align}
	J(k) = &||\hat z(k - N|k) - \bar z(k - N|x)||_{\Gamma} ^2 + ||\hat{w}_{k - N|k}^{k - 1|k}||_{\Lambda} ^2 \nonumber\\ + &||{F_N}\hat z(k - N|k) + {H_N} u_{k - N}^{k - 1} + {H_N}\hat{w}_{k - N|k}^{k - 1|k} - y_{k - N}^k||_{\Pi} ^2.\nonumber
\end{align}
The necessary condition for the minimum of the cost function $J(k)$ is given by
\begin{gather}
	\frac{{\partial J(k)}}{{\partial \hat z(k - N|k)}} = 0,\;\;
	\frac{{\partial J(k)}}{{\partial \hat{w}_{k - N|k}^{k - 1|k}}} = 0.\label{eq:necessary_condition}
\end{gather}
From (\ref{eq:necessary_condition}), the following partial differential equations can be derived
\begin{align}
	&\frac{{\partial J(k)}}{{\partial \hat z(k - N|k)}} = 2\Gamma [\hat z(k - N|k) - \bar z(k - N|x)] \nonumber\\&+  2\Pi F_N\t[{F_N}\hat z(k - N|k) + {H_N}\hat u_{k - N}^{k - 1} + {H_N}\hat{w}_{k - N|k}^{k - 1|k} - y_{k - N}^k]\nonumber\\
	&\frac{{\partial J(k)}}{{\partial \hat{w}_{k - N|k}^{k - 1|k}}} = 2\Lambda \hat{w}_{k - N|k}^{k - 1|k}\nonumber\\ &+ 2\Pi {H_N\t}[{F_N}\hat z(k - N|k) + {H_N}\hat u_{k - N}^{k - 1} + {H_N}\hat{w}_{k - N|k}^{k - 1|k} - y_{k - N}^k]. \label{eq:partial_differential_J}
\end{align}
Combining (\ref{eq:necessary_condition}) and (\ref{eq:partial_differential_J}), it gives
\begin{align}
	\hat z(k - N|k) =& {(\Gamma + \Pi {F_N\t}{F_N})^{ - 1}}
	[\Pi{F_N\t} y_{k - N}^k - \Pi{F_N\t}{H_N} u_{k - N}^{k - 1} \nonumber\\ &+ \Gamma \bar z(k - N|k) - \Pi{F_N\t}{H_N}\hat{w}_{k - N|k}^{k - 1|k}]\nonumber\\
	\hat{w}_{k - N|k}^{k - 1|k} =& {(\Lambda + \Pi {H_N\t}{H_N})^{ - 1}} [\Pi {H_N\t}y_{k - N}^k \nonumber\\ &- \Pi {H_N\t}{F_N}\hat z(k - N|k) - \Pi {H_N\t}{H_N}{u}_{k - N}^{k - 1}].\label{eq:closed_form_solution}
\end{align}
The result in (\ref{eq:solution}) can be derived from (\ref{eq:closed_form_solution}). The proof is completed. 
\eproof

The estimated uncertainties $\hat{f}$ using MH-ESO are given by
\begin{gather}
\hat{f} = D(y_1)\hat{z}_3(k-N|k)
\end{gather}
where $z_3(k-N|k)$ represents the bottom $6 + n$ elements of $\hat{z}(k-N|k)$.

{
Let $\rho_r(A)$ denote the spectral radius of a square matrix 
A. The MH-ESO is input-to-state stable, and the result in \cite{sui2014linear} can be used to prove this under the condition that $\rho_r(A_0) <1$.
}

\brem
The two equations in (\ref{eq:closed_form_solution}) can be expressed as a system of linear equations
\begin{gather}
{N_1}d_{\rm{opt}} = {N_2}\label{eq:solveLinearEquation}
\end{gather}
where $d_{\rm{opt}} = \text{col}(\hat{z}(k-N|k), \hat{w}^{k-1|k}_{k-N|k})$, 
\begin{gather}
{N_1} = \left[ {\begin{array}{*{20}{c}}
		{\Gamma + \Pi {F_N\t}{F_N}}&{\Pi {F_N\t}{H_N}}\\
		{\Pi {H_N\t}{F_N}}&{\Lambda + \Pi {H_N\t}{H_N}}
\end{array}} \right]\nonumber\\
{N_2} = \left[ {\begin{array}{*{20}{c}}
		{\Pi {F_N\t}y_{k - N}^k - \Pi {F_N\t}{H_N}u_{k - N}^{k - 1} + \Gamma \bar z(k - N|k)}\\
		{\Pi {H_N\t}[y_{k - N}^k - {H_N}u_{k - N}^{k - 1}]}
\end{array}} \right].\nonumber
\end{gather}
In practical applications, to reduce the computation time for solving Problem 1, we do not need to obtain $d_{\rm{opt}}$ using the conclusion in Theorem \ref{thm:thm4.1} by inverting matrices. Instead, we can use the sparse LU decomposition method \cite{gill1987maintaining} to solve the linear equations (\ref{eq:solveLinearEquation}).
\erem
\brem
The average computation time for the linear equation (\ref{eq:solveLinearEquation}) increases as the MHE horizon $N$ is set to a larger value. Table \ref{table:Computation_Time} presents the average computation time for the linear equation (\ref{eq:solveLinearEquation}) corresponding to $N$, when the quadruped robot with $n = 12$ degrees of freedom is trotting in place, excluding model uncertainties. The test is conducted on a single computer with an Intel i7-13700KF processor at 3.4 GHz, and each test lasts for approximately 20 seconds.
 
\begin{table}[h]
	\begin{center}
		\caption{The Average Computation Time}
		\begin{tabular}{c|c||c|c}
			\hline
			\hline
			N & Computation Time (ms)& N & Computation Time (ms)\\
			\hline
			1 & 0.168153  & 8& 5.42422 \\
			\hline
			2 & 0.461572 & 10& 8.21532 \\
			\hline
			4 & 0.966662 & 12& 19.9099 \\
			\hline
			6 & 3.12035 & 14& 27.229 \\
			\hline
		\end{tabular}\label{table:Computation_Time}
	\end{center}
\end{table}

A shorter horizon $N$ improves computational efficiency and increases uncertainty estimation bandwidth, but it is more sensitive to noise. In contrast, a longer $N$ enhances estimation smoothness and performance, though at the expense of higher computational load and reduced uncertainty estimation bandwidth. The parameter $N$ can be chosen to balance the uncertainty estimation bandwidth and the impact of noise.

\erem

\brem
The proposed MH-ESO has the advantage of suppressing the influence of measurement noise, especially when compared to the ESO, as presented in Section \ref{sec:MH_ESO_comparison}. Other methods, such as adaptive control can also be used to estimate uncertainties and integrated into the proposed three-level framework to address uncertainties in legged robots, particularly when measurement noise can be neglected in the system.
\erem

\subsection{Signal Processing}\label{sec:signal_processing}

\begin{figure}[!ht]
	\centering
	\includegraphics[scale=0.032]{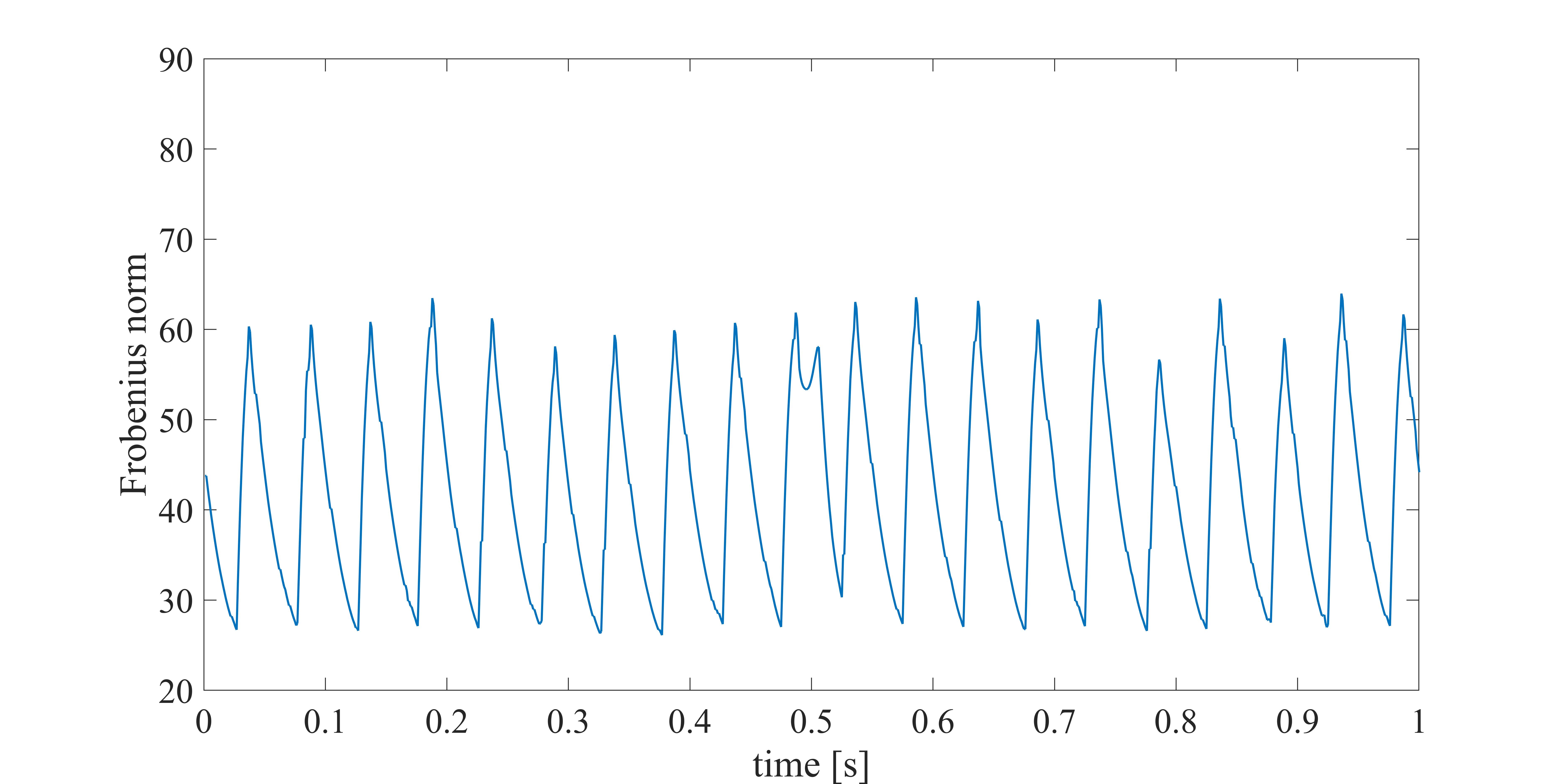}
	\caption{Frobenius norm of the uncertainties $\hat{f}$ estimated by MH-ESO when the robot, carrying a 10kg load, is standing in place.}
	\label{fig_remark}
\end{figure}

Note that the GRFs $F_{\rm{ref}}$ are used as the input to the system in (\ref{eq:ESO_1}) or (\ref{eq:ESOinMHE}), as opposed to the real robot's GRFs $F$. The discrepancy between $F$ and $F_{\rm{ref}}$ is estimated using the estimator. This discrepancy transitions from zero to non-zero, or vice versa, as the GRFs $F_{\rm{ref}}$ and $F$ transition from zero to non-zero, or vice versa. Since the discrepancy is estimated at high frequency, the estimated uncertainties undergo significant changes when the discrepancy transitions from zero to non-zero, or vice versa. To reduce noise caused by the interaction force switching, the MAF \cite{zhu2023proprioceptive} and low-pass filter \cite{sombolestan2024adaptive} can be used to mitigate the influence of interaction noise on the estimated uncertainties in legged robots. Additionally, we provide a saturator to limit the estimated uncertainties within a bounded range. 
Let 
\begin{gather}
{\hat f_{\rm{process}}} = \left\{ \begin{array}{l}
	\alpha \hat f/||\hat f||,\quad \text{if}\;\; ||\hat{f}|| > \alpha\\
	{\hat f}, \quad \quad \quad \quad \text{otherwise}
\end{array} \right.\label{eq:saturator}
\end{gather}
denote the processed estimated uncertainties produced by the saturator. The saturator only bounds the input signal without introducing signal delay, unlike the MAF and low-pass filter.
\brem
The parameter $\alpha$ can be chosen according to the Frobenius norm of the estimated uncertainties $\hat{f}$ when applied to robots standing in place without moving. For example, Fig. \ref{fig_remark} shows the Frobenius norm of the uncertainties $\hat{f}$ estimated by MH-ESO when the robot is standing in place while carrying a 10 kg load. The norm of the estimated uncertainties lies within the interval  $[25, 70.1]$, so we can choose $\alpha$ to lie within this range to handle model uncertainties due to a 10 kg load or any load less than 10 kg.
\erem

\section{Robust Model Predictive Control}\label{sec:robust_mpc}

In this section, we aim to track the reference trajectory $\dot{X}_{\rm{ref}} := [\dot{h}_{\rm{ref}}\t, \dot{L}_{\rm{ref}}\t, \dot{q}_{\rm{ref}}\t]\t$ generated by high-level control, by adjusting the GRFs of the centroidal dynamics using robust MPC.

We first calculate the change in the momentum derivatives of the legged system when uncertainties are present, compared to when the system is free of uncertainties, as follows: 
\begin{gather}
d_c = \left[ {\begin{array}{*{20}{c}}
		{\dot h}\\
		{\dot L}\\
		{\dot q}
\end{array}} \right] - \left[ {\begin{array}{*{20}{c}}
		{{{\dot h}_{{\rm{ref}}}}}\\
		{{{\dot L}_{{\rm{ref}}}}}\\
		{{{\dot q}_{{\rm{ref}}}}}
\end{array}} \right]. \label{eq:dc}
\end{gather}
The derivatives of centroidal momentum are obtained from \cite{orin2013centroidal} 
\begin{gather}
	\left[ {\begin{array}{*{20}{c}}
			{\dot h}\\
			{\dot L}
	\end{array}} \right] = {A}(q)\ddot q + {\dot A}\dot q. \label{eq:dotCM}
\end{gather}
Then, the centroidal dynamics can be rewritten as
\begin{gather}
	{\dot x_c} = \left[ {\begin{array}{*{20}{c}}
			{\dot h}\\
			{\dot L}\\
			{\dot q}
	\end{array}} \right] = 	f_h(q,\dot{q},\ddot{q}). \label{eq:dotxc}
\end{gather}
where
\begin{gather}
	{f_h}(q,\dot q,\ddot q) = \left[ {\begin{array}{*{20}{c}}
			{A(q)\ddot q + A\dot q}\\
			{\dot q}
	\end{array}} \right].
\end{gather}
Since the acceleration $\ddot{q}$ in (\ref{eq:dotxc}) cannot be measured directly, we obtain the acceleration estimation $\ddot{\hat{q}}$ as
	\begin{align}
		\ddot{\hat{q}} = D(q)^{-1}(&-C(q,\dot{q})\dot{q}-G(q) + S\t \tau \nonumber\\ &+ J(q)\t F_{\rm{ref}} + \hat{f}_{\rm{process}})\nonumber
	\end{align}
	with $\hat{f}_{\rm{process}}$ being given in (\ref{eq:saturator}). From (\ref{eq:dc}), the estimated difference in momentum derivatives, $\hat{d}_c$, is given by
	\begin{gather}
		\hat{d}_c = f_h(q,\dot{q},\ddot{\hat{q}}) - \left[ {\begin{array}{*{20}{c}}
				{{{\dot h}_{{\rm{ref}}}}}\\
				{{{\dot L}_{{\rm{ref}}}}}\\
				{{{\dot q}_{{\rm{ref}}}}}
		\end{array}} \right].
\end{gather}

Subsequently, we construct a robust MPC approach to
determine the reference trajectories for the robot’s GRFs,
while accommodating uncertainties. 
$d_c$ is considered for the compensation in the centroidal dynamics,
\begin{gather}
\dot{x}_c = f(x,u) + d_c, 
\end{gather}
to track the reference trajectory $\dot{X}_{\rm{ref}}$.
We use the estimation $\hat{d}_c$ to replace $d_c$ in the compensated centroidal dynamics for the robust MPC. The continuous dynamics (\ref{eq:centroidal_dynamics})  are discretized over intervals of the prediction horizon $[t,t + T_0]$ for the formulation of the nonlinear MPC trajectory optimization problem. Choosing the time horizon of the MPC problem as $M-1$, thus the discrete interval is set to $\delta_{t_0} = T_0/(M-1)$. Define 
\begin{gather}
J_{u}(x) = (x - X_{\rm{ref}}(t + T_0))\t Q^u_t (x - X_{\rm{ref}}(t+T_0))\nonumber
\end{gather}
and
\begin{align}
\varphi_{u,i}(x, u) =& (x - X_{\rm{ref}}(t+i\delta_{t_0}))\t Q^u_s(x - X_{\rm{ref}}(t+i\delta_{t_0})) \nonumber\\&+ (u - U_{\rm{ref}}(t+i\delta_{t_0}))\t R(u - U_{\rm{ref}}(t+i\delta_{t_0})). \nonumber
\end{align}
where $X_{\rm{ref}}$ and $U_{\rm{ref}}$ are obtained by solving the nominal MPC problem in high-level control. The nonlinear robust MPC trajectory optimization problem can be formulated as 
\begin{align}
\text{min}_{X, U}\;\;J_{u}(x_M)& + \sum ^{M-1}_{i=0} \varphi_{u,i}(x_k, u_k) + l_{i}(x_k,u_k)\nonumber\\
s.t. \;\;\;\;\;\;\; &x_{0,k} - {x}_{m,k} = 0,\nonumber\\
\;\;\;\;\; &x_{i+1,k} - x_{i,k} - f_c(x_{i,k},u_{i,k})\delta t_0 ={ \hat{\textbf{\textit{d}}_\textbf{c}}\delta t_0}, \nonumber\\
\;\;\;\;\; &g_i(x_{i,k},u_{i,k}) = 0,\;\; i = 0,\cdots,N-1. \label{eq:MPCproblemWithUncertainty}
\end{align}
where $f_c(x_{i,k},u_{i,k})$ and the constraint function $g_i(x_{i,k},u_{i,k})$ is given in (\ref{eq:MPC_problem}). The difference between the MPC problem in (\ref{eq:MPC_problem}) and the MPC problem in (\ref{eq:MPCproblemWithUncertainty}) is that $\hat{d}_c$ is introduced in (\ref{eq:MPCproblemWithUncertainty}), but it is not included in the MPC problem in (\ref{eq:MPC_problem}).

The solution to the MPC problem (\ref{eq:MPCproblemWithUncertainty}) is denoted as $X = X^{u}_{\rm{ref}}: = [(h^u_{\rm{ref}})\t, (q^u_{\rm{ref}})\t]\t$ and $U = U^u_{\rm{ref}}: = [(F^u_{\rm{ref}})\t,(\dot{q}^{u,j}_{\rm{ref}})\t]\t$,  corresponding to discrete time $[t,t+\delta_{t_0},\cdots,t+M\delta_{t_0}]\t$. The reference velocity is given by
\begin{gather}
	\dot{q}^{u}_{\rm{ref}} = \left[ {\begin{array}{*{20}{c}}
			{\dot{q}_{\rm{ref}}^{u,b}}\\
			{\dot{q}_{\rm{ref}}^{u,j}}
	\end{array}} \right].
\end{gather} 
The  reference velocity $\dot{q}^{u,b}_{\rm{ref}}$ for the base can be obtained as
\begin{gather}
	\dot{q}^{u,b}_{\rm{ref}} = A^{-1}_b(q^{u}_{\rm{ref}})\left(\left[ {\begin{array}{*{20}{c}}
			h^{u}_{\rm{ref}}\\
			{L}^{u}_{\rm{ref}}
	\end{array}} \right] - {A_j}({q^{u}_{\rm{ref}}}){\dot q^{u,j}_{\rm{ref}}}\right)
\end{gather}
where $A(q)$ is partitioned as $ A(q):=[A_b(q),A_j(q)]  $ with $A_b(q) \in \mathbb{R}^{6 \times 6}$. 
Partition the matrix $D(q)$ in (\ref{eq:dynamics})  as  
\begin{gather}
	D(q) = \left[ {\begin{array}{*{20}{c}}
			{{D_{11}}(q)}&{{D_{12}}(q)}\\
			{{D_{21}}(q)}&{{D_{22}}(q)}
	\end{array}} \right]
\end{gather}
where $D_{11}(q) \in \mathbb{R}^{6 \times 6}$. 
The reference acceleration vector for the actuated joints ${\ddot q}^{u,j}_{\rm{ref}} \in \mathbb{R}^{n}$ is obtained by directly differentiating the   joint velocity $\dot{q}^{u,j}_{\rm{ref}}$. The  base reference acceleration $\ddot{q}^{u,b}_{\rm{ref}}$ for the base can be calculated as follows
\begin{align}
	\ddot{q}^{u,b}_{\rm{ref}} =& D_{11}(q^{u}_{\rm{ref}})^{-1}(S_{\theta}(-C(\dot{q}^{u}_{\rm{ref}},q^{u}_{\rm{ref}})\dot{q}^{u}_{\rm{ref}} -G(q^{u}_{\rm{ref}}) + J\t F^{u}_{\rm{ref}})\nonumber\\ &- D_{12}{(q^{u}_{ref})}{\ddot q}^{u,j}_{\rm{ref}}).\nonumber
\end{align}
where   $S_{\theta} = [I_6, 0_{6 \times n}]$.
The reference acceleration is given by
\begin{gather}
	{\ddot q^{u}_{\rm{ref}}} = \left[ {\begin{array}{*{20}{c}}
			{{{\ddot q}^{u,b}}_{\rm{ref}}}\\
			{{{\ddot q}^{u,j}}_{\rm{ref}}}
	\end{array}} \right].\label{eq:ddotqref}
\end{gather}

\section{Low-level Control}\label{sec:low_level_control}

Similar to \cite{11079307}, we use the hierarchical WBC to address uncertainties in the legged system. The detailed formulation of WBC is presented in \cite{bellicoso2016perception}. In WBC, each task is formulated with equality and inequality constraints on generalized accelerations, joint torques, and GRFs. Each task can be defined as a set of linear equality and inequality constraints. The hierarchical WBC can be formulated as 
\begin{align}
\text{find}\;\;&\ddot{q}_d, F_d, \tau_d\nonumber\\
\text{s.t.}\;\;\;\; &\underline{u} \le u \le \overline{u} \quad \quad \quad \quad \quad \quad \;\;\quad \quad \quad\;\;\; \text{(Task Torque limits)}\nonumber\\
&D(q)\ddot{q}_d + C(q,\dot{q})\dot{q} + G(q) = S\t \tau_d + J\t F_d + \hat{\textbf{\textit{f}}}_\textbf{process}  \;\nonumber\\\nonumber &\quad \quad \quad \quad  \quad \quad \quad \quad \quad \quad \quad \quad 
\quad \quad \quad \quad  \text{(Task  Dynamics)}\nonumber\\
&J\ddot{q}_d + \dot{J}\dot{q} = 0\quad \quad \quad \quad \quad \;\; \quad  \text{(Task  No contact motion)}\nonumber\\
&C_fF_d \le 0 \quad \quad \quad \quad \quad \quad \quad \quad \quad \quad  \;\; \text{(Task Friction cone)} \nonumber\\
&S_{w}F_d = 0\quad \quad \quad \quad \quad \;\quad   \quad \; \text{(Task  No swing foot force)}\nonumber\\
&\ddot{q}_d - \ddot{q}^u_{\rm{ref}} = 0 \quad \quad \quad \quad \quad  \;\;\; \text{(Task Acceleration tracking)} \nonumber\\
&F_d - F^{u}_{\rm{ref}} = 0 \;\; \quad \quad \quad  \quad   \text{(Task  Foot end force tracking)}\nonumber
\end{align}
where $\underline{u}$ and $\overline{u}$ represent the lower and upper limits of the torque, $q$ and $\dot{q}$ are obtained by the feedback of the robot system, and $J$ and $\dot{J}$ are computed based on the gait and feedback. $S_w$ is the swing leg selection matrix. Tasks are solved in a strictly prioritized order. The tasks used in the paper are described in Table \ref{table:task}. 

\begin{table}[h]
\begin{center}
	\caption{Each Task Is Associated With A Priority}
	\begin{tabular}{l|l}
		\hline
		\hline
		Priority & Task\\
		\hline
		1 & Dynamics + Friction cone + No swing foot force \\&+ Torque limits	\\
		\hline
		2 & Acceleration tracking\\
		\hline
		3 & No contact motion + Foot end force tracking\\
		\hline
	\end{tabular}\label{table:task}
\end{center}
\end{table}
Finally, the actuation joint torque $u_d$ can be computed using the PD control law as follows: 
\begin{gather}
	u_d = \tau^*_d + K_p(q^{u,j}_{\rm{ref}} - q^j) + K_d(\dot{q}^{u,j}_{\rm{ref}} - \dot{q}^j)
\end{gather}
where $\tau^*_d$ is obtained by solving WBC problem, $K_p \in \mathbb{R}^{n \times n}$ and $K_d \in \mathbb{R}^{n \times n}$ are positive definite diagonal matrices of proportional and derivative gains, respectively. $q^{u,j}_{\rm{ref}}$ and $\dot{q}^{u,j}_{\rm{ref}}$ represent the bottom $n$ elements of $q^u_{\rm{ref}}$ and $\dot{q}^u_{\rm{ref}}$, respectively.

\section{Simulation Verification}\label{sec:simulation_comparison}
To evaluate the effectiveness and versatility of the T-WB-DRC, we conduct a series of simulation experiments on legged robots. First, we present the advantages of the MH-ESO over the ESO. Second, we compare the performance of the proposed T-WB-DRC with the WB-DRC introduced in \cite{11079307} and the widely adopted standard whole-body control framework (standard WB-C) from \cite{grandia2023perceptive}, while model uncertainties, external disturbances, and fault tolerance are considered. 
Both the standard WB-C and WB-DRC employ a two-level control framework, comprising high-level and low-level controls, with identical high-level control mechanisms. The primary distinction between the WB-DRC and the standard WB-C lies in the integration of an adaptive ESO within the WB-DRC, which estimates uncertainties in the low-level control, while the standard WB-C does not address such uncertainties. In contrast to both the standard WB-C and WB-DRC, the T-WB-DRC adopts a three-level control framework. It retains the same high-level control as the WB-C and WB-DRC, while incorporating the same low-level control as the WB-DRC.

All simulations are performed on a single PC (Intel i7-13700KF, 3.4 GHz). For the simulations, the control system is implemented using ROS Noetic in the Gazebo simulator. The simulations are carried out on both a humanoid robot (the upper body of the humanoid robot is modeled as a rigid body), which has $n = 12$ degrees of actuation freedom, and the Unitree A1 quadruped robot, which also has $n = 12$ degrees of actuation freedom. For the T-WB-DRC, the high-level control, MH-ESO, robust MPC in mid-level control, and low-level control operate in four distinct threads. In contrast, for the WB-DRC and standard WB-C, the high-level control and low-level control are executed in two different threads. We select the same operating frequency for high-level control across all three methods, as well as the same operating frequency for low-level control.

For the T-WB-DRC, both the nominal and robust MPC operate at specific frequencies and run on separate threads from the WBC. When the MPC and WBC operate simultaneously, the robust MPC uses the last reference trajectory calculated from the most recent nominal MPC solution, while the WBC uses the last reference trajectory calculated from the most recent robust MPC solution. Once the MPC optimization problem is solved, the reference trajectory is updated. We employ a read-write lock to facilitate data transfer between the high-level control and mid-level control, as well as between the mid-level control and the WBC in the low-level control.

\subsection{Impact of Noise on ESO and MH-ESO} \label{sec:MH_ESO_comparison}
To show the MH-ESO and ESO sensitivity to noise, we introduced Gaussian random noise $\mathcal{N}(0,0.001)$ to each element in the real position $q^{\rm{real}}$ to simulate authentic running environments as
\begin{gather}
q = q^{\rm{real}} + \mathcal{N}(0,0.001).\label{eq:gaussian_noise}
\end{gather} 
We conduct a simulation on the robot carrying a 10 kg load and standing in place and compare the Mean Squared Error (MSE) and the variance of the Frobenius norm of the estimated uncertainties $\hat{f}$ using MH-ESO and ESO. This comparison demonstrates the impact of noise on the estimated uncertainties using both MH-ESO and ESO while the robot remains stationary with the 10 kg load.
The MSE and variance of the Frobenius norm of the estimated uncertainties are defined as
\begin{gather}
MSE = \frac{1}{N}\sum\limits_{i = 1}^N {{{({{\hat y}_i} - {\bar{y}_i})}^2},\quad Var = } \frac{1}{N}\sum\limits_{i = 1}^N {{{({{\hat y}_i} - \rho )}^2}}
\end{gather}
where ${{\hat y}_i}$ and $\bar{y}_i$ represent the norms of the estimated uncertainties obtained using position feedback with noise and the average norms of the estimated uncertainties obtained using position feedback without noise, respectively, for MH-ESO or ESO; $\rho$ represents the sample mean norms of the estimated uncertainties obtained using position feedback with noise. 

\begin{table}[h]
	\begin{center}
		\caption{MH-ESO and ESO Setting}
		\begin{tabular}{l|l|l}
			\hline
			\hline
			Parameter & Value for MH-ESO & Value for ESO\\
			\hline
			$\omega_0$ & 200 & 200	\\
			\hline
			$N$ & 4 & $\times$\\
			\hline
			$\Gamma$ &  $10 \times I_{54 \times 54}$& $\times$\\
			\hline
			$\Lambda$ &  $1 \times I_{54N \times 54N}$& $\times$\\
			\hline
			$\Pi$ &  $0.01$&$\times$\\
			\hline
		\end{tabular}\label{table:MH-ESO_setting}
	\end{center}
\end{table}

\begin{figure}[!ht]
	\centering
	\begin{minipage}{1.1\linewidth}
		\includegraphics[width=0.85\linewidth]{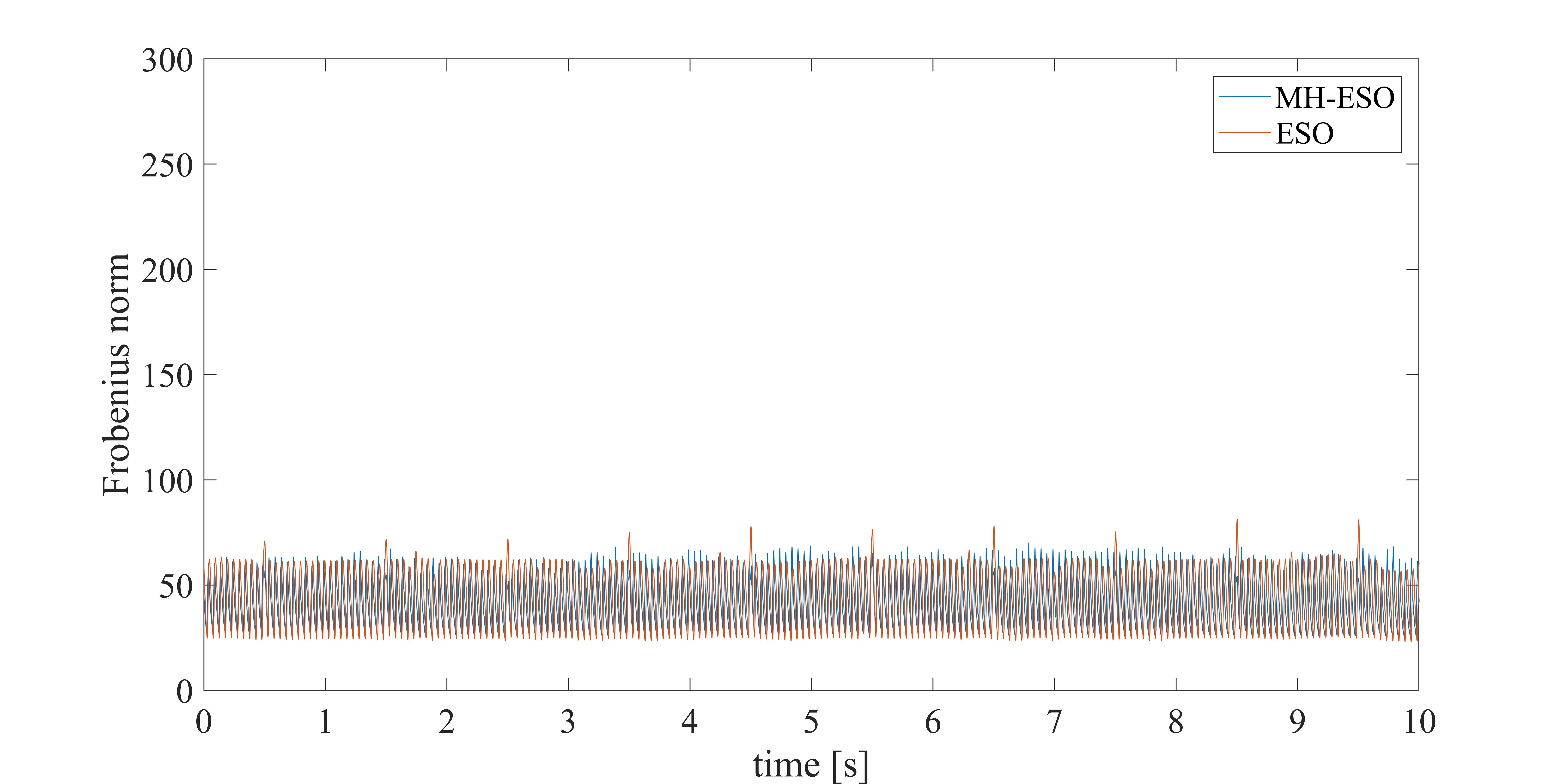}
	\end{minipage}
	\begin{minipage}{1.1\linewidth}
		\includegraphics[width=0.85\linewidth]{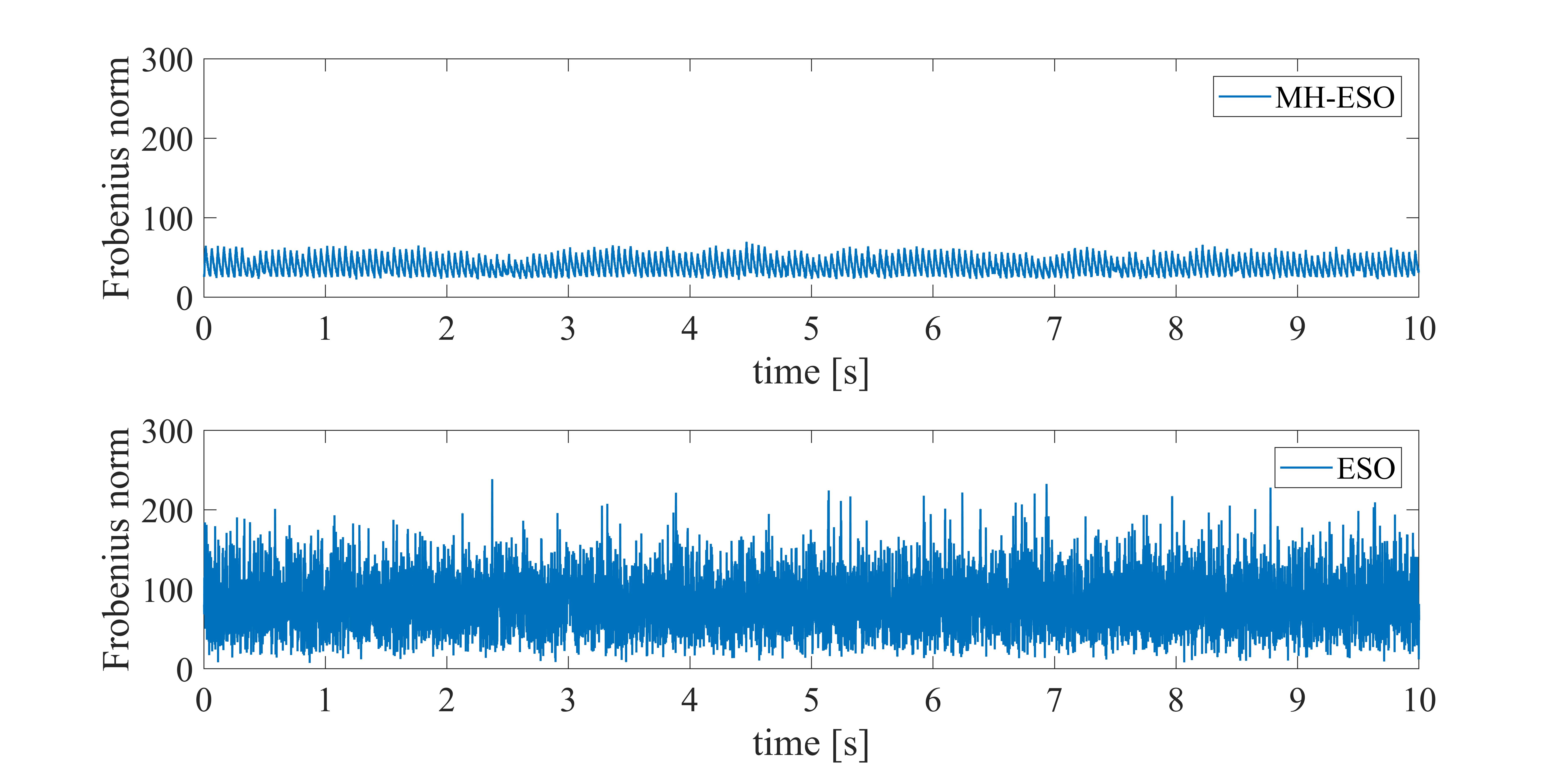}
	\end{minipage}
	\caption{The Frobenius norms of the uncertainties estimated by MH-ESO or ESO.}
	\label{fig_noise}
\end{figure}

The MH-ESO and ESO settings are presented in Table \ref{table:MH-ESO_setting}.  The top figure of Fig. \ref{fig_noise} shows that the Frobenius norms of the uncertainties estimated by MH-ESO and ESO are similar when the robot's position feedback is free from Gaussian random noise.  The bottom figure of Fig. \ref{fig_noise} shows the Frobenius norms of the estimated uncertainties for both methods, suggesting that MH-ESO is less sensitive to noise than ESO. The average MSE and variance results, presented in Table \ref{table:MSE_Variance}, indicate that MH-ESO enhances noise suppression capability compared to ESO when both observers are set to the same bandwidth.

\begin{table}[h]
	\begin{center}
		\caption{MSE and Variance Results}
		\begin{tabular}{l|l|l}
			\hline
			\hline
			Method & MSE & Var\\
			\hline
			MH-ESO & 95.127 & 90.0257	\\
			\hline
			ESO & 2754.4& 1190.3 \\
			\hline
		\end{tabular}\label{table:MSE_Variance}
	\end{center}
\end{table}

\subsection{Simulations on Quadruped Robots}
The operating frequency for high-level control is set to 20 Hz with the prediction horizon $T = 1 s$ and the total number of steps $H = 50$, while the low-level control operates at 1000 Hz. The robust MPC also operates at 20 Hz with the prediction horizon $T_0 = 1 s$ and the total number of steps $M = 50$. 
We define the MPC optimization problem solving
time as the duration between the start of the MPC calculation and its completion. Note that the MPC
optimization problem solving time needs to be less than 50 ms at a frequency of 20 Hz for the
simulation comparison. Each step of the MPC optimization must be solved within 50 ms.

\begin{figure}[!ht]
	\centering
	\begin{minipage}{0.30\linewidth}
		\centering
		\includegraphics[width=0.9\linewidth]{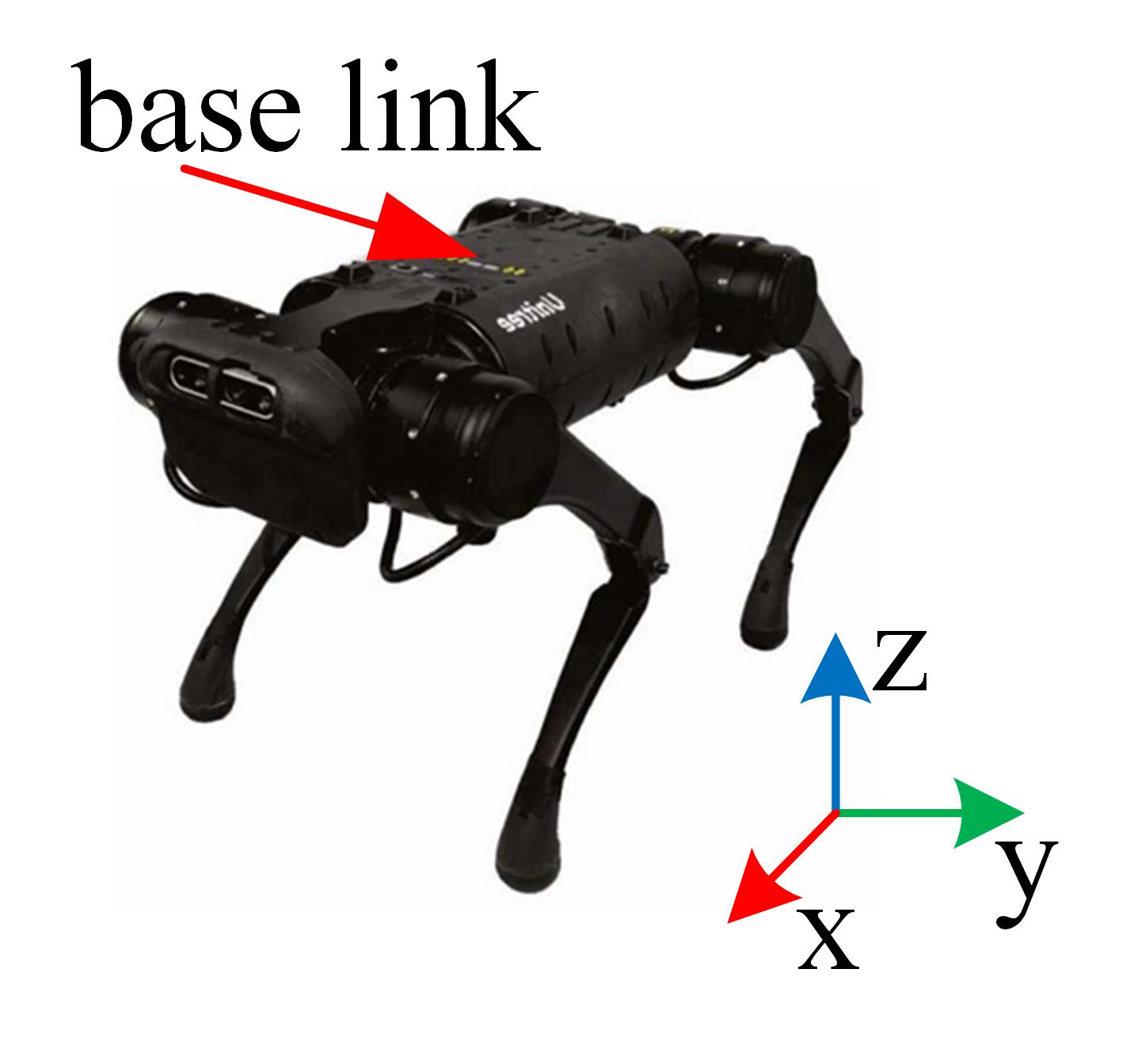}
		\label{chutian1}
	\end{minipage}
	\begin{minipage}{0.56\linewidth}
		\centering
		\includegraphics[width=0.9\linewidth]{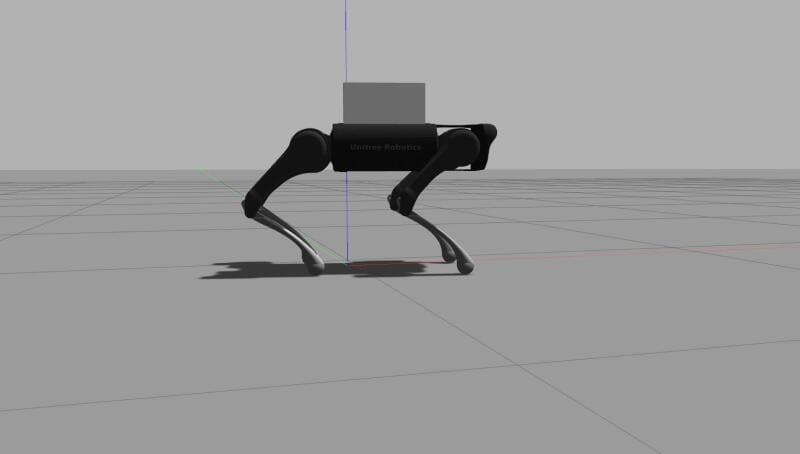}
		\label{chutian2}
	\end{minipage}
	\caption{The Unitree A1 robot (left) and its model in the simulation environment with a $10$ kg load (right).}
	\label{A1_gazebo}
\end{figure}

We select MH-ESO as the disturbance estimator to estimate uncertainties and use the saturator as the signal processing module, with the parameter $\alpha = 25$ set in the saturator. The operating frequency for MH-ESO is set to 1000 Hz, and the configuration of MH-ESO is provided in Table \ref{table:MH-ESO_setting}. The desired base link (see Fig. \ref{A1_gazebo}) height is set to 0.31 m and the desired roll angles of the base link is set to 0 rad. To simulate realistic running environments, we introduced Gaussian random noise 
$\mathcal{N}(0,0.001)$ to each element of the real position $q^{\rm{real}}$, as shown in (\ref{eq:gaussian_noise}).

\begin{figure}[!ht]
	\centering
	\includegraphics[scale=0.032]{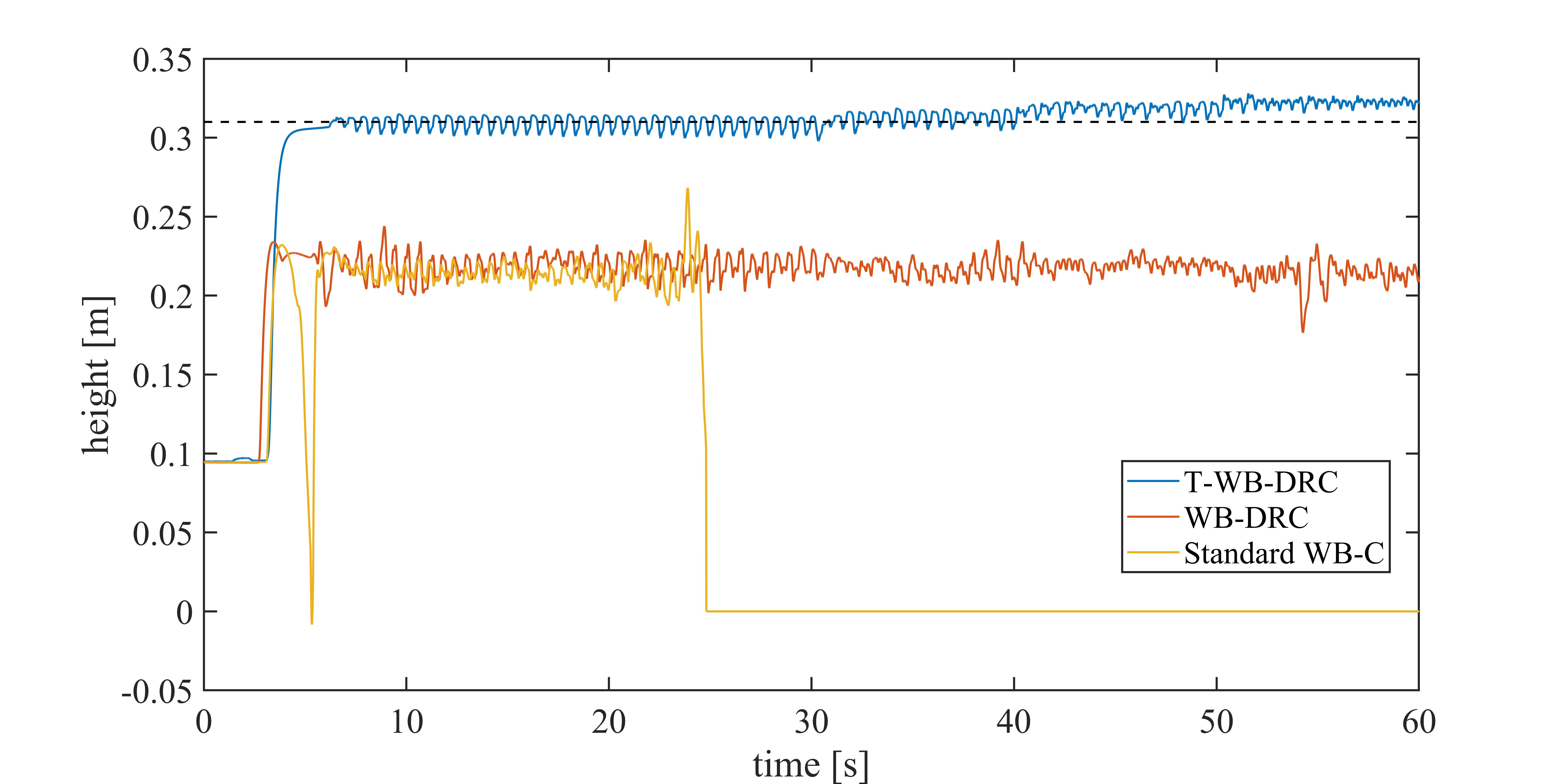}
	\caption{The height of the base link of the robot when the robot moves carrying a 10 kg load.}
	\label{fig_model_uncertainties}
\end{figure}

\begin{figure}[!ht]
	\centering
	\begin{minipage}{1.1\linewidth}
		\includegraphics[width=0.85\linewidth]{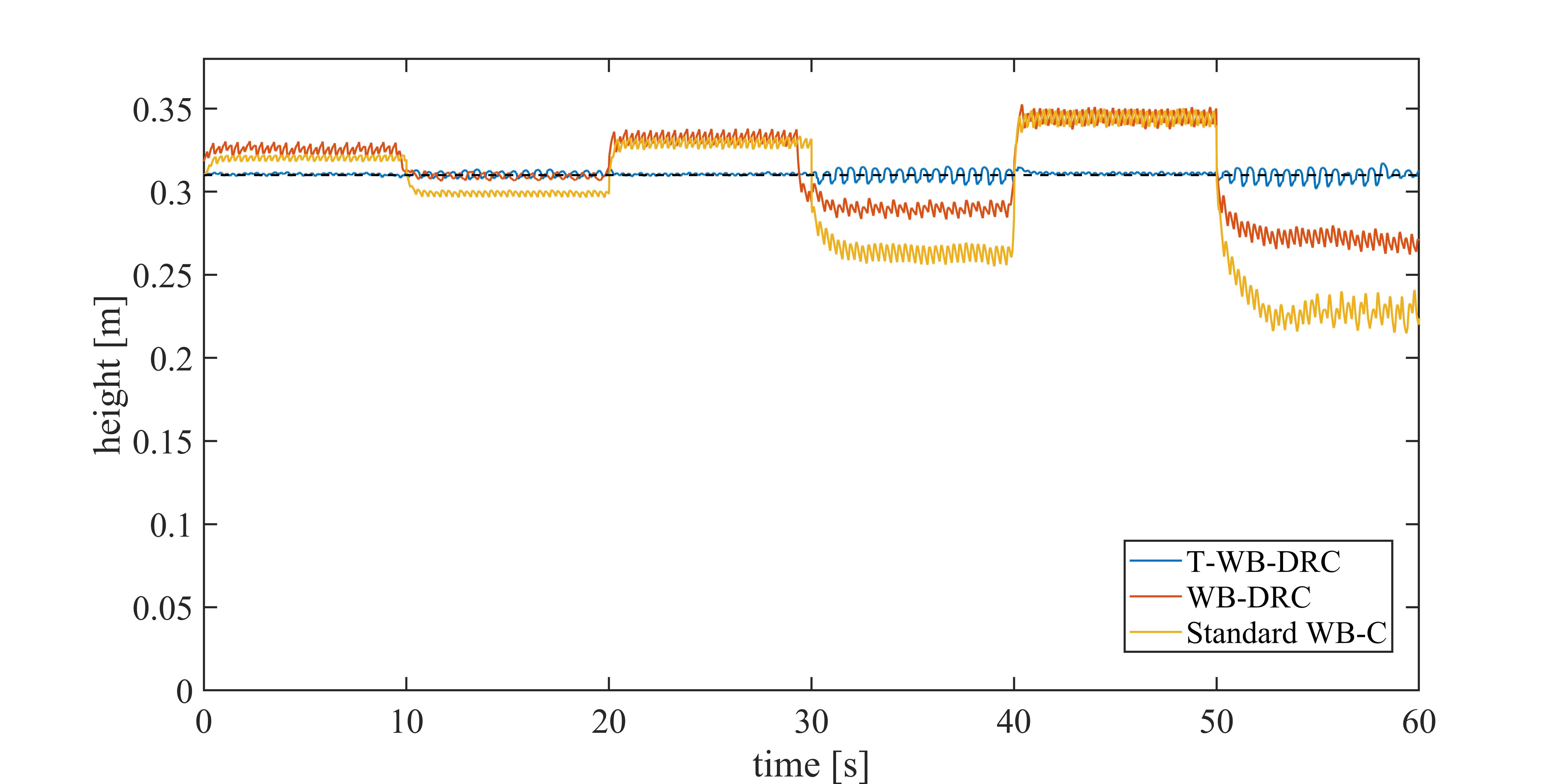}
	\end{minipage}
	\begin{minipage}{1.1\linewidth}
		\includegraphics[width=0.85\linewidth]{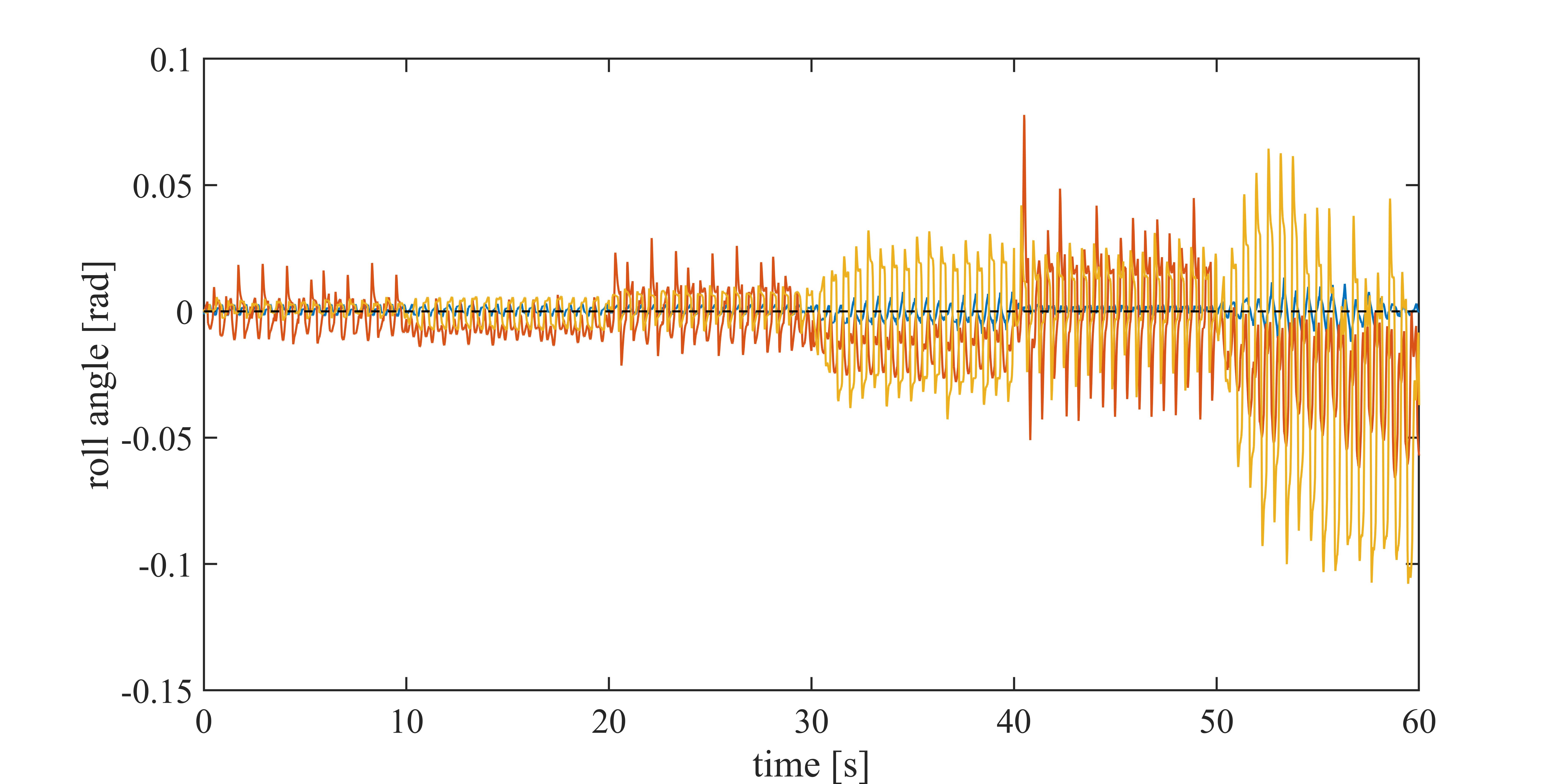}
	\end{minipage}
	\caption{The height and roll angle of the robot's base link when constant forces are applied to it.}
	\label{fig_external_diturbance}
\end{figure}

\begin{figure}[!ht]
	\centering
	\includegraphics[scale=0.032]{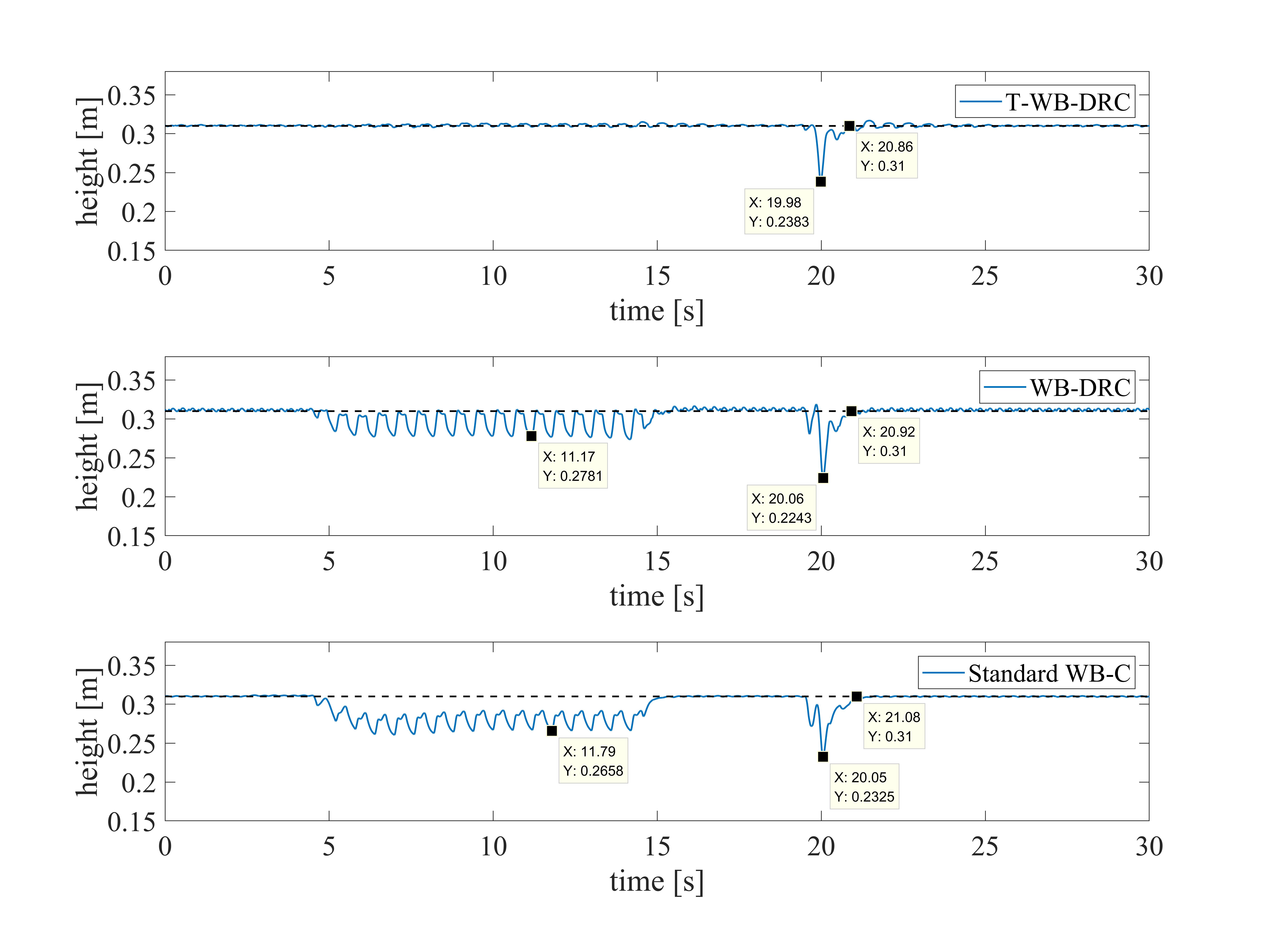}
	\caption{The height of the robot's base link when the knee joint motor output torque is cut.}
	\label{fig_fault}
\end{figure}

To demonstrate the effectiveness and versatility of T-WB-DRC, we conduct three simulation comparisons: (1) under model uncertainties with the robot carrying a 10 kg load, (2) under external disturbances with a constant force applied to the base link, and (3) to assess fault tolerance by reducing motor output torques. Fig. \ref{fig_model_uncertainties} illustrates the height of the base link over time as the robot, carrying a 10 kg load, walks at different velocities along the $x$-axis. The robot is commanded to move on the flat ground with velocities of $-0.1$ m/s, $0.1$ m/s, $0.2$ m/s, $0.4$ m/s, and $0.6$ m/s from 10 s to 60 s, each for a duration of 10 seconds.  The results demonstrate that the robot using T-WB-DRC maintains the body height closer to 0.31 m, showcasing its ability to effectively handle model uncertainties. Fig. \ref{fig_external_diturbance} shows the
height of the base link from the ground and the roll angle of the base link over time under the influence of various constant external forces. External forces of $20$ N, $-20$ N, $40$ N, $-60$ N, $80$ N, and $80$ N are applied to the base link along the
$z$-axis, each for a duration of 10 seconds. The results demonstrate that the proposed T-WB-DRC compensates for these external disturbances better than the WB-DRC and the standard WB-C.
Furthermore, we intentionally reduce the output torque of the robot's knee joint on the right back leg by 50\% from 5 s to 10 s and by 90\% from 19.8 s to 20 s to demonstrate the fault tolerance capability. The base link height of the robot while stepping in place is shown in Fig. \ref{fig_fault} for T-WB-DRC, WB-DRC, and standard WB-C. For T-WB-DRC, under long-term motor output torque faults (from 5 s to 10 s), the height of the base link closely follows the desired trajectories, outperforming the robot using the standard WB-C and WB-DRC. In the case of short-term motor output torque faults (from 19.8 s to 20 s), the base link height of the robot using T-WB-DRC tracks the desired trajectories more rapidly compared to both the robot using the WB-DRC and the robot using standard WB-C. These results show that the
robot using the WB-DRC exhibits better fault tolerance and enhanced
robustness compared to the robot using the standard WBC.

Table \ref{table:Sum_differences} presents the MSE of the humanoid robot's base link height relative to the desired base link height, considering the robot's response to model uncertainties over the 10 s to 40 s interval, external disturbances from 0 s to 30 s, and fault tolerance during the 0 s to 30 s interval, as observed in the previous three comparison experiments. The results demonstrate that the robot utilizing the T-WB-DRC method outperforms those using the other two methods.

\begin{table}[h]
	\begin{center}
		\caption{ MSE of Actual Height Relative to Desired Height}
		\scalebox{0.95}{
		\begin{tabular}{l|l|l|l}
			\hline
			\hline
			&\multicolumn{2}{c}{MSE}&\\
			\hline 
			Method & Model Uncertainties & External Disturbances & Faults\\
			\hline
			WB-C & 0.0627 & 0.0017	& 3.6961e-04\\
			\hline
			WB-DRC & 0.0103& 6.1195e-04 & 2.4302e-04\\
			\hline
			T-WB-DRC & 0.0025& 5.4849e-06& 2.7885e-05 \\
			\hline
		\end{tabular}}\label{table:Sum_differences}
	\end{center}
\end{table}

\subsection{Simulations on Humanoid Robots}
\begin{figure}[!ht]
	\centering
	\begin{minipage}{0.38\linewidth}
		\includegraphics[width=0.85\linewidth]{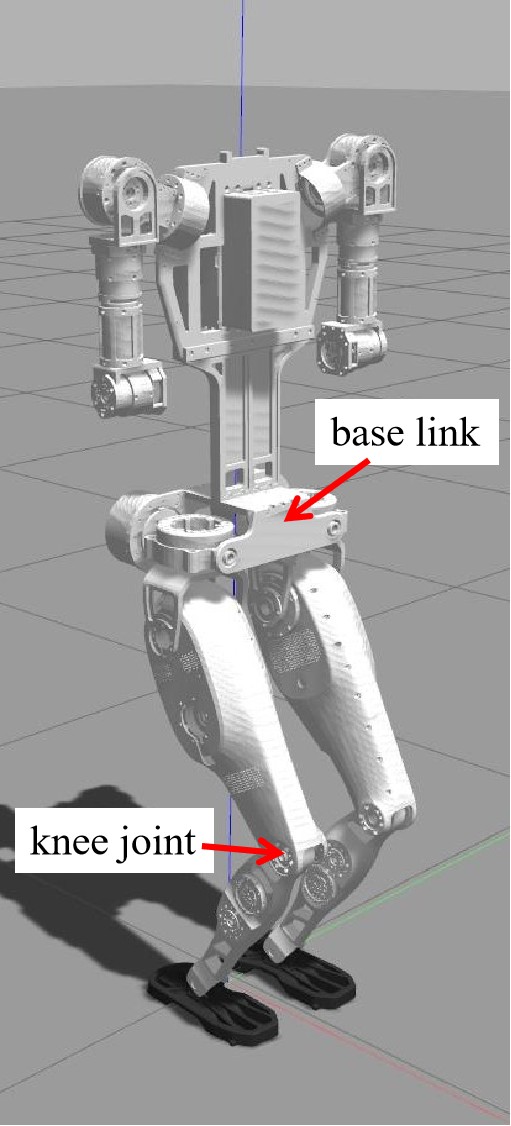}
		\label{fig1}
	\end{minipage}
	\begin{minipage}{0.48\linewidth}
		\includegraphics[width=0.85\linewidth]{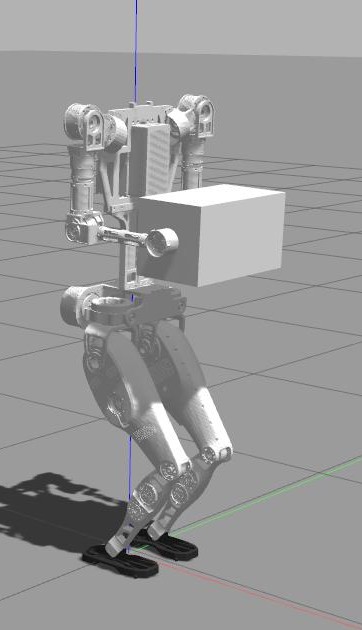}
		\label{fig2}
	\end{minipage}
	\caption{The humanoid robot in the simulation environment (left) and the humanoid robot carrying a 8 kg load (right).}
	\label{humanoid_robot}
\end{figure}

The operating frequency for high-level control is set to 100 Hz with the prediction horizon $T = 1s$ and the total number of steps $H = 34$, while the low-level control operates at 1000 Hz. The robust MPC also operates at 100 Hz with the prediction horizon $T_0 = 1 s$ and the total number of steps $M = 34$. The operating frequency for disturbances estimator is set to 1000 Hz. The desired base link height (see Fig. \ref{humanoid_robot}) is set to 0.80 m.

\begin{figure}[!ht]
	\centering
	\includegraphics[scale=0.032]{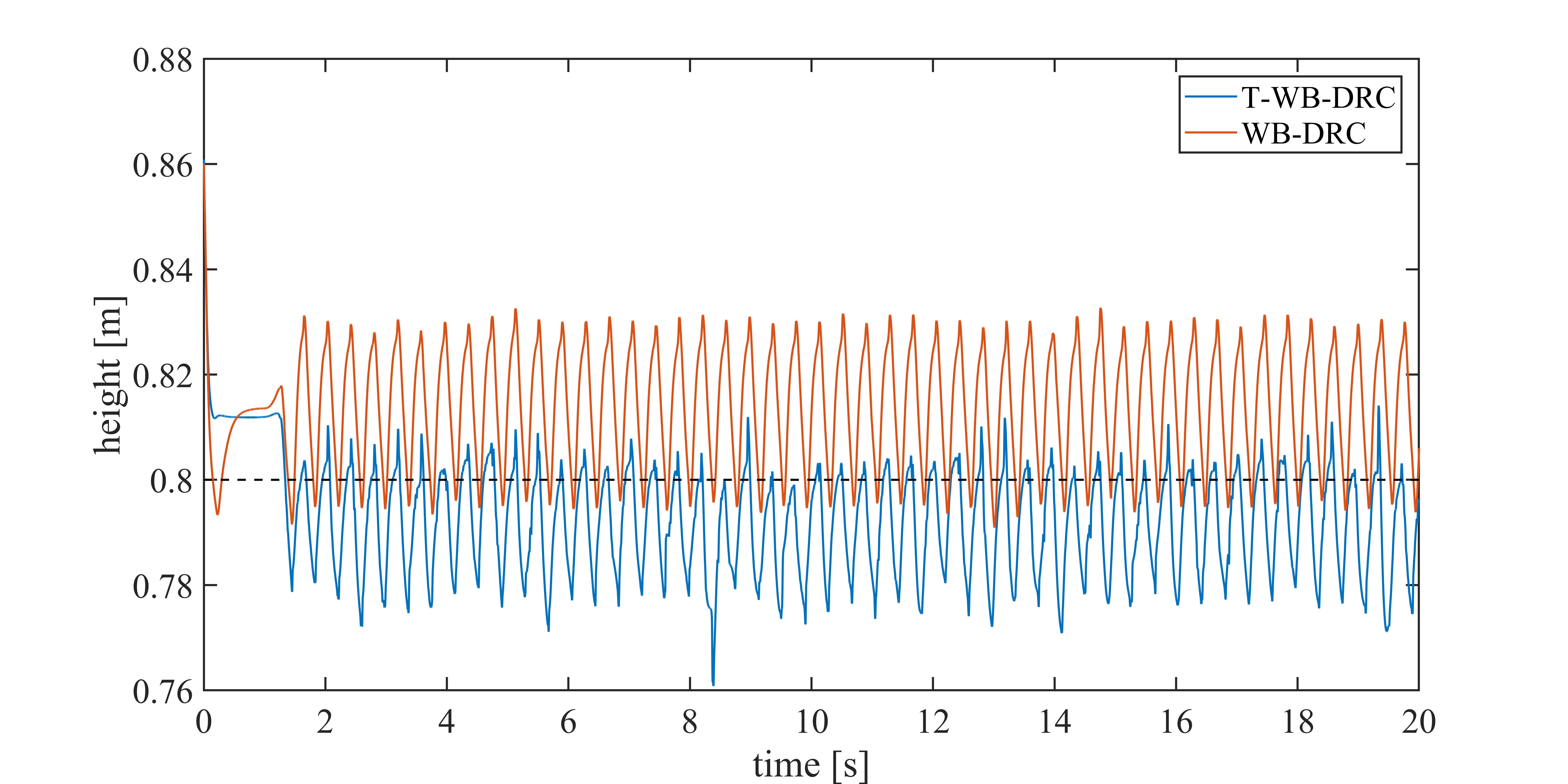}
	\caption{The height of the humanoid robot's base link when the knee joint motor output torque is reduced by 30\% with $\omega_0 = 100$.}
	\label{fig_humanoid_100}
\end{figure}

\begin{figure}[!ht]
	\centering
	\includegraphics[scale=0.032]{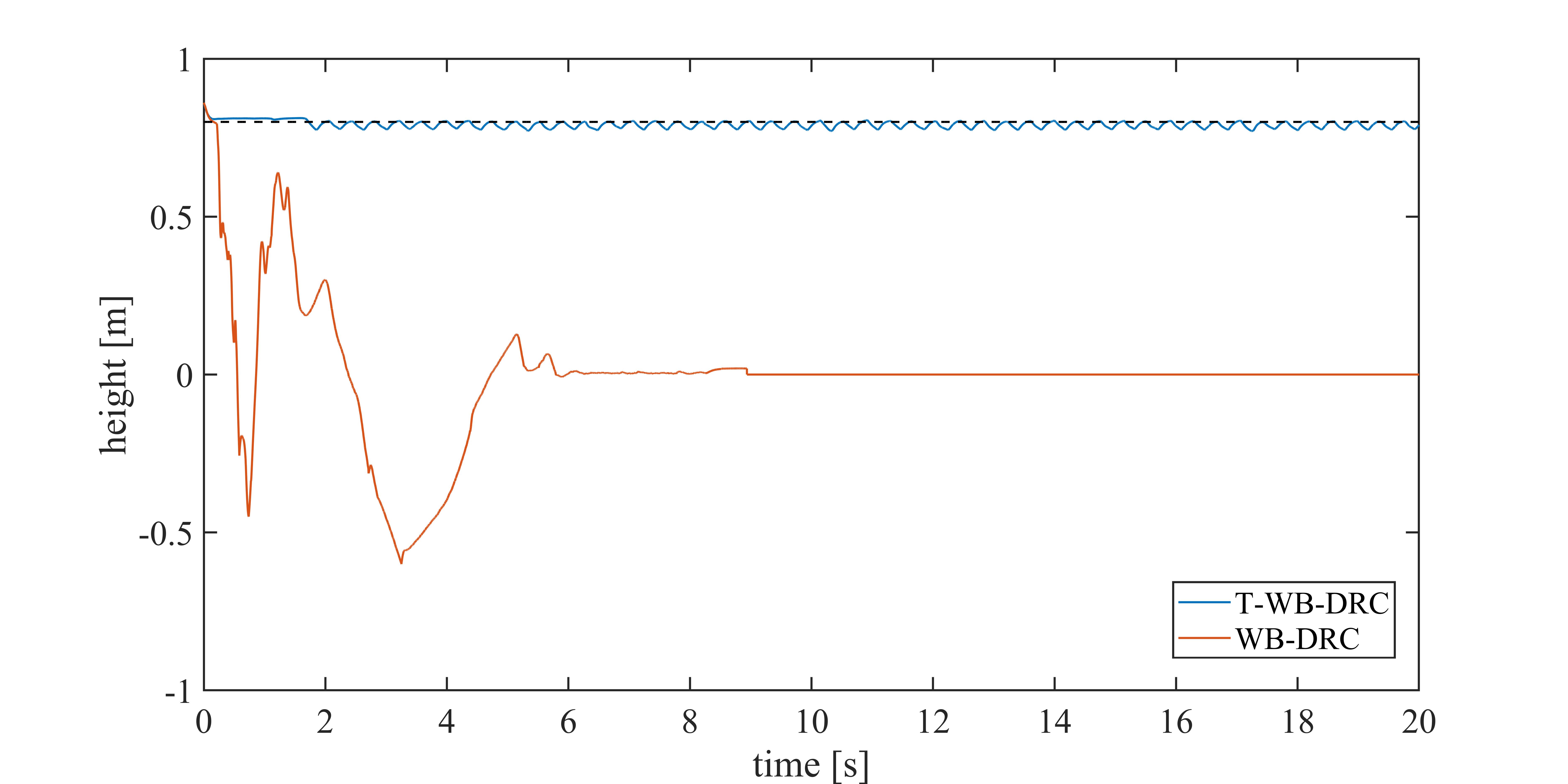}
	\caption{The height of the humanoid robot's base link when the knee joint motor output torque is reduced by 30\% with $\omega_0 = 300$.}
	\label{fig_fault_humanoid}
\end{figure}

To demonstrate the fault tolerance capability of the humanoid robot, we intentionally reduce the output torque of the right leg's knee joint  (see Fig. \ref{humanoid_robot}) by 30\%. We select MH-ESO as the disturbance estimator to estimate uncertainties and use the MAF as the signal processing module, with the moving horizon window as 3. Fig. \ref{fig_humanoid_100} and \ref{fig_fault_humanoid} show the height of the humanoid robot's base link when the knee joint motor's output torque is reduced by 30\%, with the ESO bandwidth set to $\omega_0 = 100$ and $\omega_0 = 300$, respectively. Table \ref{table:MSE_humanoid} shows the MSE of the humanoid robot's base link height relative to the desired base link height. These results illustrate that the T-WB-DRC has the better fault tolerance capability than WB-DRC when the ESO bandwidth $\omega_0 = 100$. As shown in Fig. \ref{fig_fault_humanoid}, the robot falls at the start of controller operation due to the increased noise influence as the ESO bandwidth is increased. This demonstrates that the T-WB-DRC has the potential to achieve better locomotion performance than WB-DRC, particularly in the presence of noise.

\begin{table}[h]
	\begin{center}
		\caption{MSE of Actual Height Relative to Desired Height}
		\begin{tabular}{l|l|l}
			\hline
			\hline
			Method & MSE ($\omega_0$ = 100) & MSE ($\omega_0$ = 300)\\
			\hline
			MH-ESO & 1.6673e-04 & 1.8417e-04	\\
			\hline
			ESO & 3.2156e-04& $\times$ \\
			\hline
		\end{tabular}\label{table:MSE_humanoid}
	\end{center}
\end{table}

\begin{figure}[!ht]
	\centering
	\includegraphics[scale=0.032]{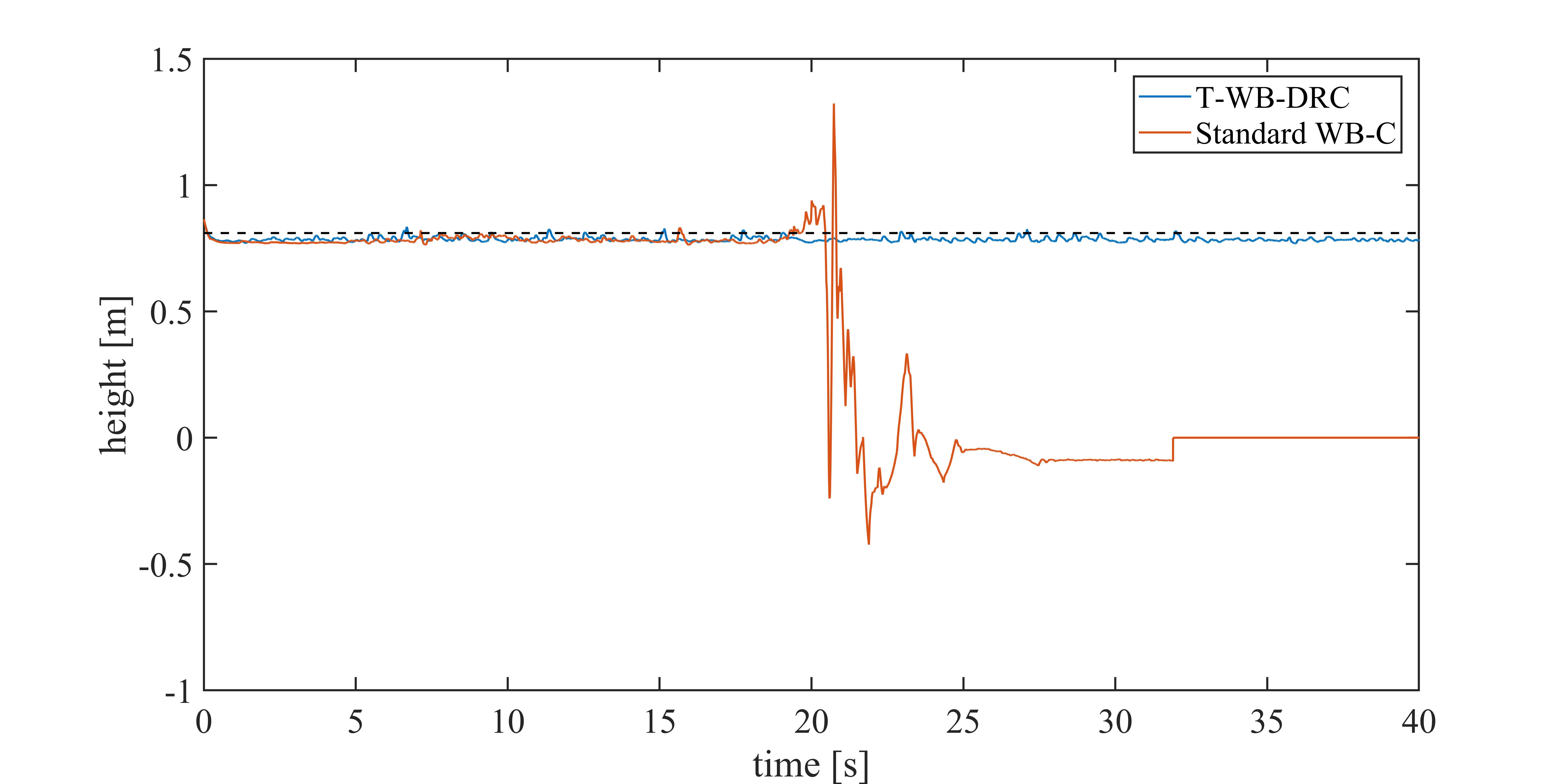}
	\caption{The height of the humanoid robot's base link when the robot holds an 8 kg load.}
	\label{fig_humanoid_load}
\end{figure}

To demonstrate the ability to handle model uncertainties, we conduct a simulation comparison in which the humanoid robot holds an 8 kg load while trotting in place, as shown in Fig. \ref{humanoid_robot}. We select ESO as the disturbance estimator to estimate uncertainties and use the MAF as the signal processing module, with the moving horizon window as 7. Fig. \ref{fig_humanoid_load} shows the base link height of the humanoid robot while holding an 8 kg load and trotting in place. It demonstrates that the robot using T-WB-DRC trots for the full 40 s, whereas the robot using standard WB-C falls after approximately 20 s.

\section{Experimental Verification}\label{sec:experimental_verification}
\begin{figure}[!ht]
	\centering
	\includegraphics[scale=0.18]{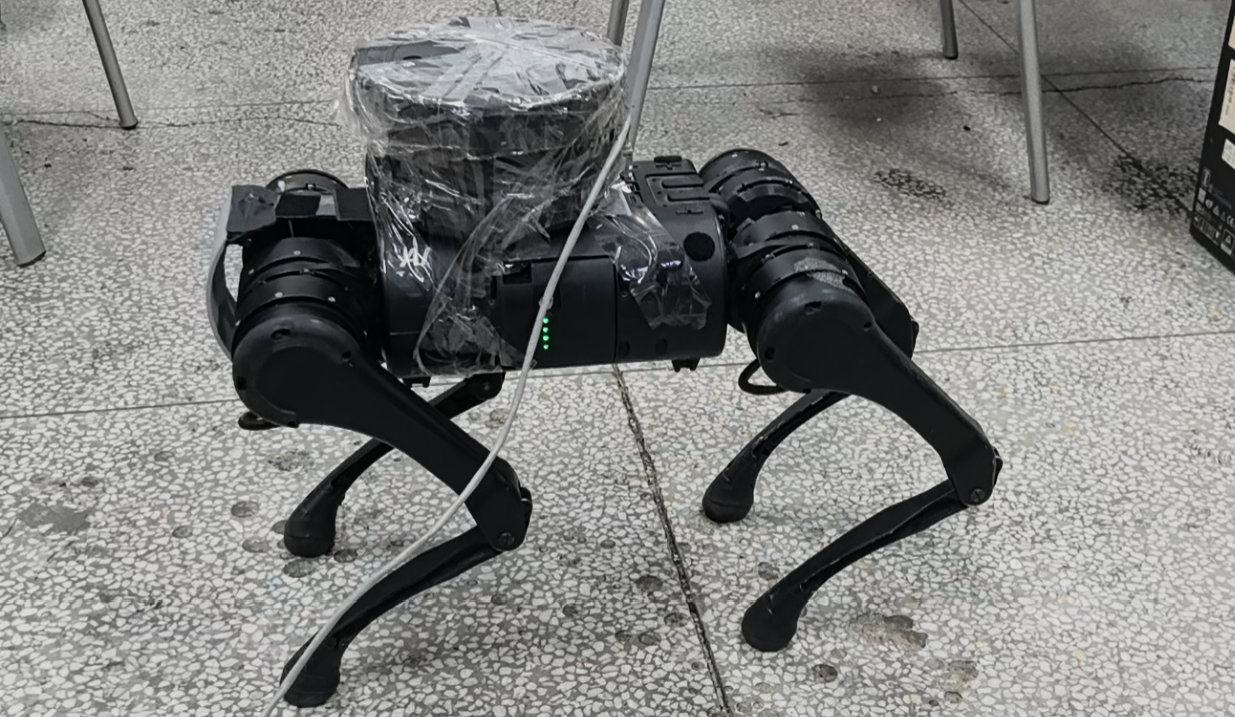}
	\caption{The Unitree A1 robot with 5 kg load.}
	\label{fig_A1}
\end{figure}

In this section, we first present two real-robot comparison experiments conducted on a single PC (Intel i7-13700KF, 3.4 GHz) using the Unitree A1 quadruped robot. The robot is equipped with joint encoders featuring a 15-bit resolution and an IMU operating at a 1 kHz sampling frequency, all performed on flat ground. These experiments highlight the advantages of the proposed T-WB-DRC over both the WB-DRC and the standard WB-C.
We then perform a real-robot experiment on challenging terrain outdoors using a low-performance hardware platform, a microcomputer with an Intel i7-1165G7, 2.8 GHz. First, we reduce the knee joint output torque by 50\% to evaluate the fault tolerance of the robot using three different methods. Next, the robot trots in place with a 5 kg load, as shown in Fig. \ref{fig_A1}, to assess its robustness to model uncertainties. Additionally, to demonstrate the control framework’s ability to handle model uncertainties and external disturbances on rough terrain, we conduct experiments with the robot trotting on such terrain while carrying the 5 kg load, highlighting the T-WB-DRC’s low dependence on hardware. 
The operating frequency for high-level control is set to 40 Hz with the prediction horizon $T = 1$ s and the total number of steps $H = 34$, while the low-level control operates at 1000 Hz with the prediction horizon $T_0 = 1$ s and the total number of steps $M = 34$. The robust MPC also operates at 40 Hz with the prediction horizon $T = 1 s$ and discrete interval $\delta_t = 0.03 s$. We choose MH-ESO as the disturbances estimator and the saturator as the signal processing module. The desired base link height is set to 0.31 m and the desired roll angle of the base link is set to 0 rad.

\begin{figure}[!ht]
	\centering
	\begin{minipage}{1.1\linewidth}
		\includegraphics[width=0.85\linewidth]{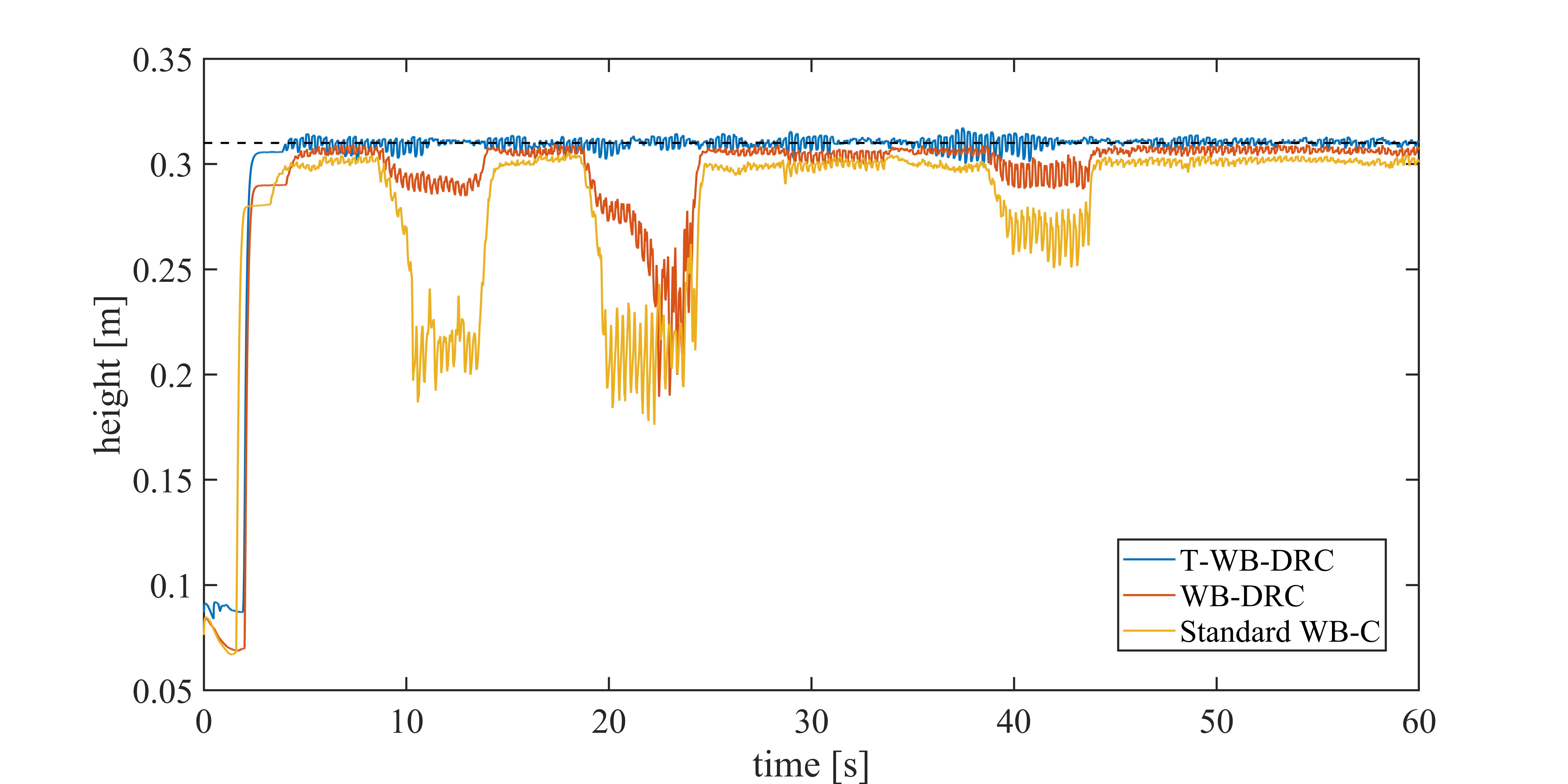}
	\end{minipage}
	\begin{minipage}{1.1\linewidth}
		\includegraphics[width=0.85\linewidth]{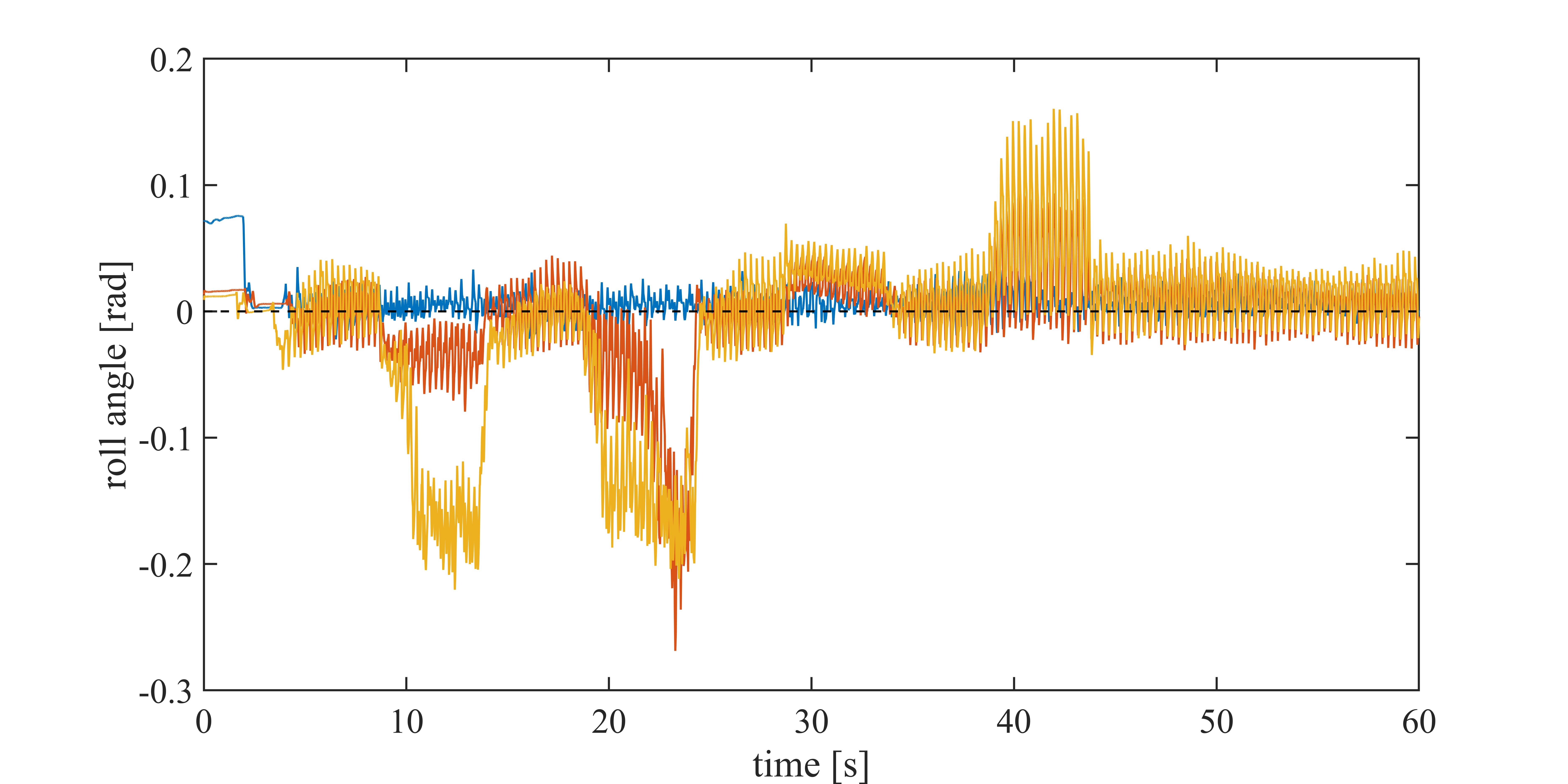}
	\end{minipage}
	\caption{The height and roll angle of the robot's base link when the output torque of the knee joints is reduced by 50\%.}
	\label{fig_falut_tolerance}
\end{figure}

\begin{figure}[!ht]
	\centering
	\begin{minipage}{1.1\linewidth}
		\includegraphics[width=0.85\linewidth]{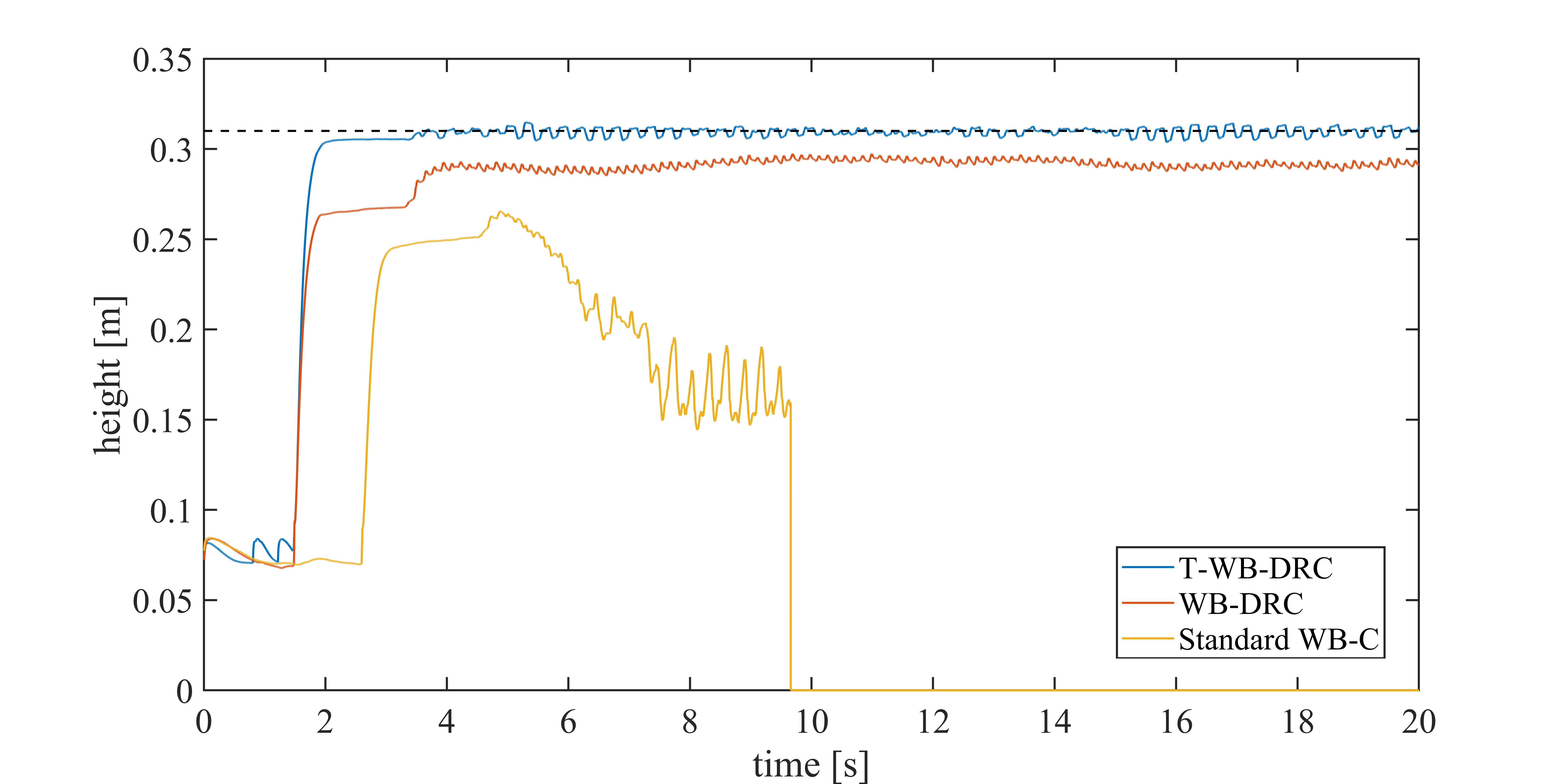}
	\end{minipage}
	\begin{minipage}{1.1\linewidth}
		\includegraphics[width=0.85\linewidth]{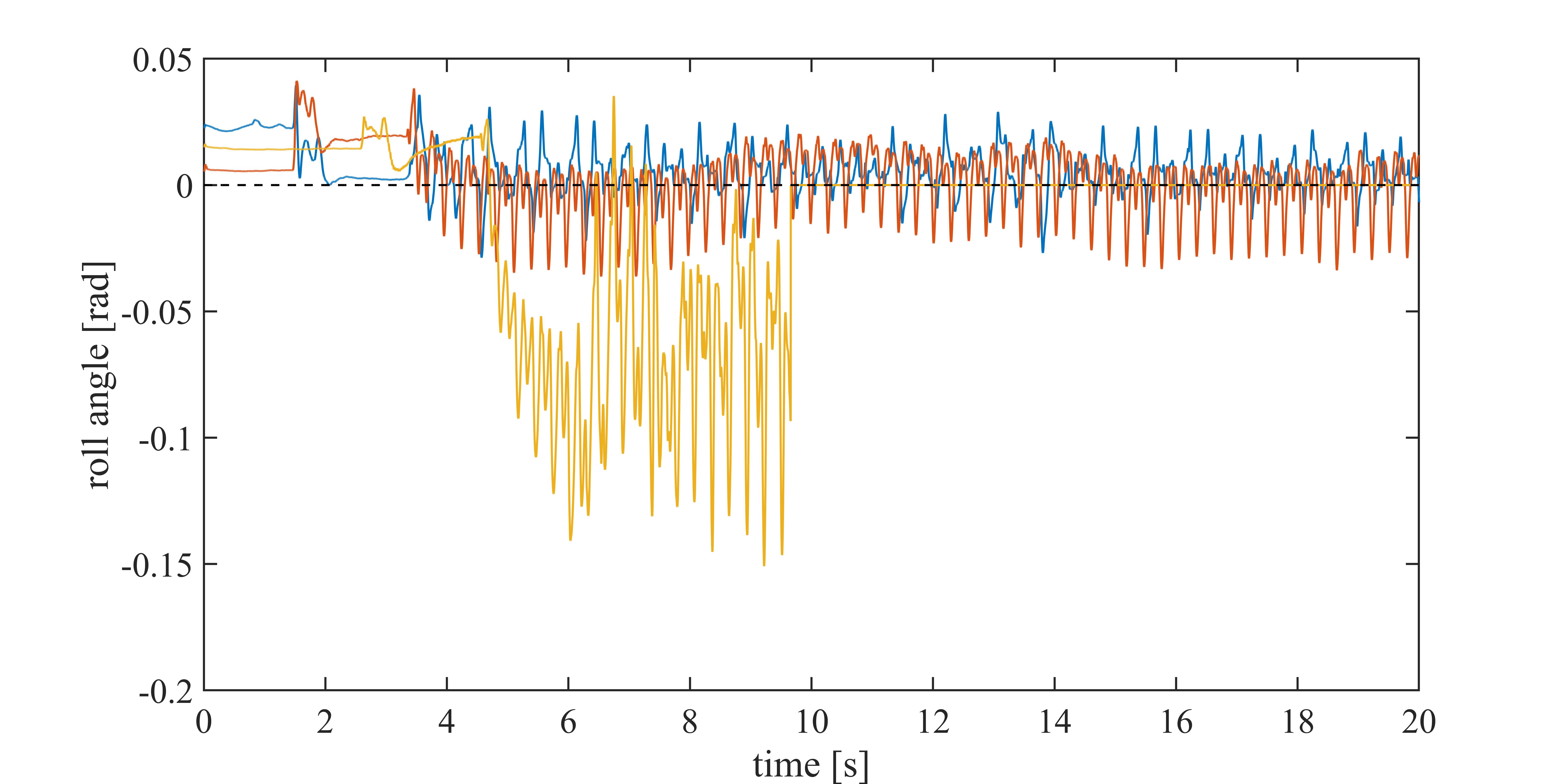}
	\end{minipage}
	\caption{The height and roll angle of the robot's base link while the robot trots in place with a 10 kg load.}
	\label{fig_carry_load}
\end{figure}

During the time intervals of 8.5s to 13.5s, 18.5s to 23.5s, 28.5s to 33.5s, and 38.5s to 43.5s, the output torque of one knee joint in each of the four legs is intentionally reduced by 50\%, with a distinct knee joint subjected to torque reduction in each interval. Fig. \ref{fig_falut_tolerance} shows the height and roll angle of the robot's base link when the output torque of the four knee joints corresponding to the four legs is reduced by 50\% for three different methods. As illustrated in Fig. \ref{fig_falut_tolerance}, the robot employing T-WB-DRC demonstrates superior fault tolerance compared to both the robot using WB-DRC and the one utilizing standard WB-C. Fig. \ref{fig_carry_load} illustrates the height and roll angles of the robot's base link while the robot trots in place, carrying a 5 kg load. As shown in Fig. \ref{fig_carry_load}, the height and roll angles of the robot's base link using T-WB-DRC closely follow the desired trajectory, while the robot using the standard WB-C experiences a failure at nearly 10 seconds. These results indicate that T-WB-DRC demonstrates superior robustness to model uncertainties compared to both WB-DRC and standard WB-C.  
Table \ref{table:exp_differences} presents the MSE of the humanoid robot's base link height relative to the desired height, considering the robot's response to model uncertainties over the 5 s to 60 s interval and external disturbances from 5 s to 20 s, as observed in the previous two comparison experiments. The results indicate that the robot employing the T-WB-DRC method outperforms those using the other two methods.

\begin{table}[h]
		\begin{center}
			\caption{ MSE of Actual Height Relative to Desired Height}
			\begin{tabular}{l|l|l}
				\hline
				\hline
				&\multicolumn{2}{c}{MSE}\\
				\hline 
				Method & Model Uncertainties & External Disturbances\\
				\hline
				WB-C & 0.0707 & 0.0017\\
				\hline
				WB-DRC & 3.4190e-04& 3.5434e-04 \\
				\hline
				T-WB-DRC &5.3580e-06& 6.4525e-06\\
				\hline
			\end{tabular}\label{table:exp_differences}
		\end{center}
\end{table}

\begin{figure}[!ht]
	\centering
	\includegraphics[scale=0.60]{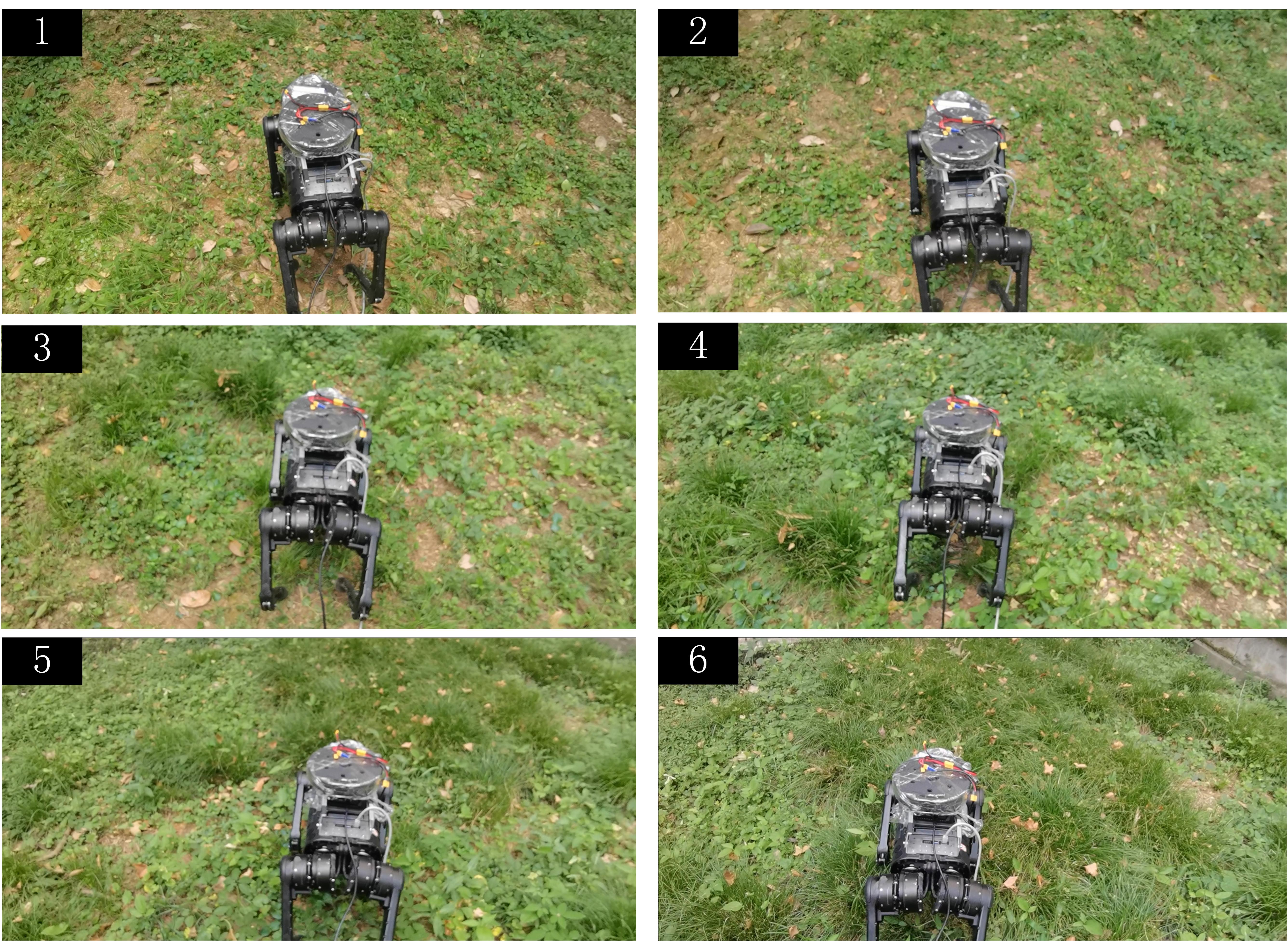}
	\caption{The robot moves on tough terrain while carrying a 5 kg load.}
	\label{fig_terrain}
\end{figure}

To demonstrate the control framework's capability in handling model uncertainties and external disturbances on rough terrain, we conduct a real-robot experiment on challenging terrain. As shown in Fig. \ref{fig_terrain}, the robot successfully navigates the rough terrain at a velocity of 0.2 m/s along the $x$-direction while carrying a 5 kg load, highlighting the effectiveness of the proposed framework in enabling the robot to traverse difficult environments.

\brem
Due to the introduction of the uncertainty estimator in WB-DRC, this method demonstrates improved performance over the standard WB-C in terms of disturbance rejection and fault tolerance. However, it also exhibits higher noise sensitivity compared to standard WB-C. By incorporating the MH-ESO, which reduces noise sensitivity relative to the standard ESO, the T-WB-DRC achieves lower noise sensitivity than WB-DRC. As a result, a larger bandwidth can be selected for the T-WB-DRC compared to the WB-DRC, leading to better disturbance rejection and fault tolerance in T-WB-DRC. A summary of the advantages and disadvantages of the three methods is provided in Table \ref{table:characteristic_comparison}. The simulation results in Section \ref{sec:simulation_comparison} and experimental results in Section \ref{sec:experimental_verification} verify the explanations above.

\begin{table*}[h]
		\begin{center}
			\caption{The Characteristic Comparison for the Three Methods}
			\centering
			\begin{tabular}{c|c|c|c}
				\hline
				\hline
				&\multicolumn{3}{c}{Method}\\
				\hline 
				Characteristic & Standard WB-C & WB-DRC & T-WB-DRC\\
				\hline
				 Disturbance Rejection& Basic  & Superior to WB-C, but inferior to T-WB-DRC	&Best \\
				\hline
				 Fault Tolerance & Basic & Superior to WB-C, but inferior to T-WB-DRC & Best\\
				\hline
				 Noise Sensitivity & Low & High & Low \\
				\hline
			\end{tabular}\label{table:characteristic_comparison}
		\end{center}
\end{table*}
\erem

\section{Conclusion}\label{sec:conclusion}
This paper presents an advanced control framework aimed at enhancing the stability and robustness of legged robots in the presence of various uncertainties, such as model uncertainties, external disturbances, and faults. The proposed framework integrates an MH-ESO to effectively estimate uncertainties and mitigate noise, ensuring accurate disturbance compensation. The introduction of the T-WB-DRC significantly improves robot performance by considering both uncertainty-free and uncertainty-affected dynamics, offering better payload transportation, disturbance rejection, and fault tolerance. Simulations on humanoid and quadruped robots demonstrate the framework's versatility, while experimental trials on a quadruped robot validate its robustness and stability under a wide range of disturbance conditions. Overall, the proposed T-WB-DRC offers a promising approach to improving the performance of legged robots in dynamic and unpredictable environments.

\appendix
\prooflater{Theorem \ref{thm}}
Let us consider the following control Lyapunov candidate
function:
\begin{gather}
	V(\eta) = \eta\t P \eta \label{eq:V}
\end{gather}
where the matrix $P$ satisfies $A\t P + PA = -I$.
Therefore, its time derivative will be
\begin{gather}
	\dot V(\eta) = -{\omega _0}{\eta \t}\eta  + \frac{{{{\tilde u}\t}{g_e\t}{C_1\t} P\eta  + {\eta\t}P{C_1}{g_e}\tilde u}}{{{\omega _0}}} \nonumber\\+ \frac{{{h_x\t}{C_2}\t P\eta  + {\eta\t}P{C_2}h_x}}{{{\omega _0}^2}}.
\end{gather}
It follows that
\begin{gather}
	\dot V(\eta ) \le  - {\omega _0}||\eta |{|^2} + \left(\frac{{2{u_b}||{C_1\t}P||}}{{{\omega _0}}} + \frac{{2{h_b}||{C_2\t}P||}}{{{\omega _0}^2}}\right)||\eta ||.\label{eq:dotVeta}
\end{gather}
From (\ref{eq:V}), we have
\begin{gather}
	V(\eta) \le \lambda_{\max}(P)||\eta||^2. \label{eq:Veta}
\end{gather}
Let $\lambda = {\omega_0}/{2\lambda_{\max}(P)}$. Combining (\ref{eq:dotVeta}) and (\ref{eq:Veta}), we have
\begin{gather}
	\dot{V}(\eta) + \lambda V(\eta) \le a(||\eta|| + \frac{b}{2})^2 + \lambda\delta_V \label{eq:dotV}
\end{gather}  
where 
\begin{gather}
	a = -\omega_0 + \lambda\lambda_{\max}(P), \; b = 2\left(\frac{{{u_b}||{C_1\t}P||}}{{{\omega _0}}} + \frac{{{h_b}||{C_2\t}P||}}{{{\omega _0}^2}}\right)/a,\nonumber\\
	\delta_V = \frac{{{{({\omega _0}{u_b}||{C_1}^TP|| + {h_b}||{C_2}^TP||)}^2}}}{{a\lambda {\omega _0}^4}}.\nonumber
\end{gather}
From (\ref{eq:dotV}), we have
\begin{gather}
	\dot V(\eta ) + \lambda V(\eta ) \le \lambda {\delta _V}.\label{eq:dotV_V}
\end{gather}
Invoking the comparison lemma \cite{khalil2002nonlinear} for (\ref{eq:dotV_V}) yields
\begin{gather}
	V(\eta) \le V(\eta_0)\exp(-\lambda t) + \delta_V(1-\exp(-\lambda t))
\end{gather}
where $\eta_0$ is the initial value of $\eta$.

Since $V(\eta) \ge \lambda_{\min}(P)||\eta||^2$, with $\lambda_{\min}(P)$ being the minimum eigenvalue of matrix $P$, it follows that
\begin{gather}
	||\eta|| \le \sqrt{\frac{ V(\eta_0)\exp(-\lambda t) + \delta_V(1-\exp(-\lambda t))}{\lambda_{\min}(P)}}.
\end{gather}
Thus, the estimation error $\eta$ is bounded. 
\eproof

\bibliographystyle{IEEEtran}
\bibliography{IEEEabrv,IEEEexample}

\vspace{-1.5cm}
\begin{IEEEbiography}[{\includegraphics[width=1in,height=1.25in,clip,keepaspectratio]{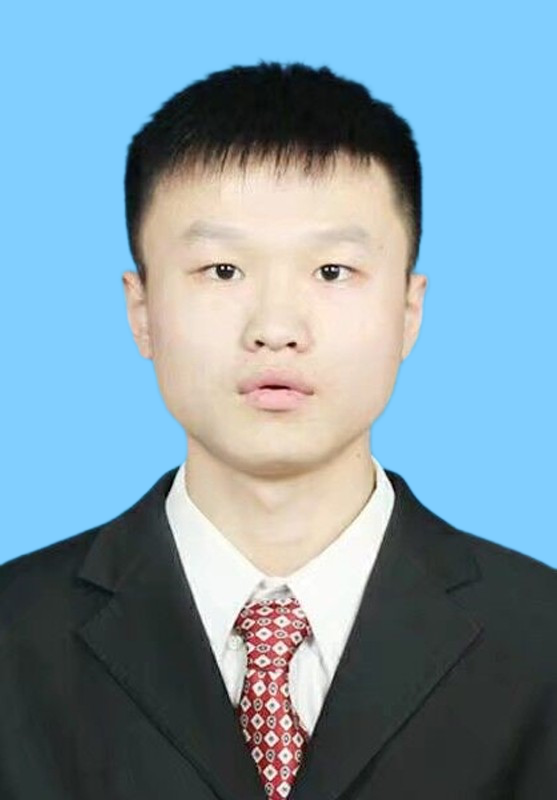}}]
	{Bolin Li} received the M.E. degree in Mechatronic Engineering from Huazhong University of Science and Technology, Wuhan, China, in 2021, and is currently pursuing the Ph.D. degree at the School of Artificial Intelligence and Automation of Huazhong University of Science and Technology. His research focuses primarily on model predictive control, active disturbance rejection control, and reinforcement learning, with applications dedicated to motor control and legged robot motion control.
\end{IEEEbiography}

\vspace{-1.5cm}
\begin{IEEEbiography}[{\includegraphics[width=1in, height=1.25in, clip, keepaspectratio]{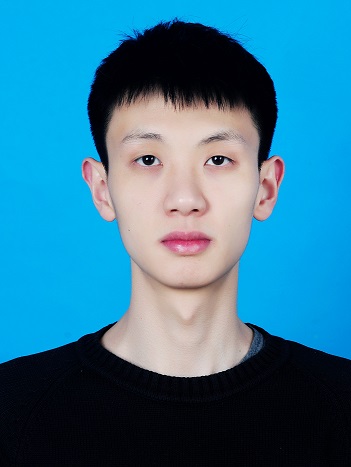}}]{Gewei Zuo}
	received the M.E. degree in control theory and engineering from Chongqing University, Chongqing, China, in 2022. He is currently pursuing the Ph.D. degree in control science and engineering with the School of Artificial Intelligence and Automation at Huazhong University of Science and Technology, Wuhan, Hubei, China.
	His research interests include nonlinear system control theory, distributed cooperative control, and distributed convex optimization.
\end{IEEEbiography}

\vspace{-1.5cm}
\begin{IEEEbiography}[{\includegraphics[width=1in, height=1.25in, clip, keepaspectratio]{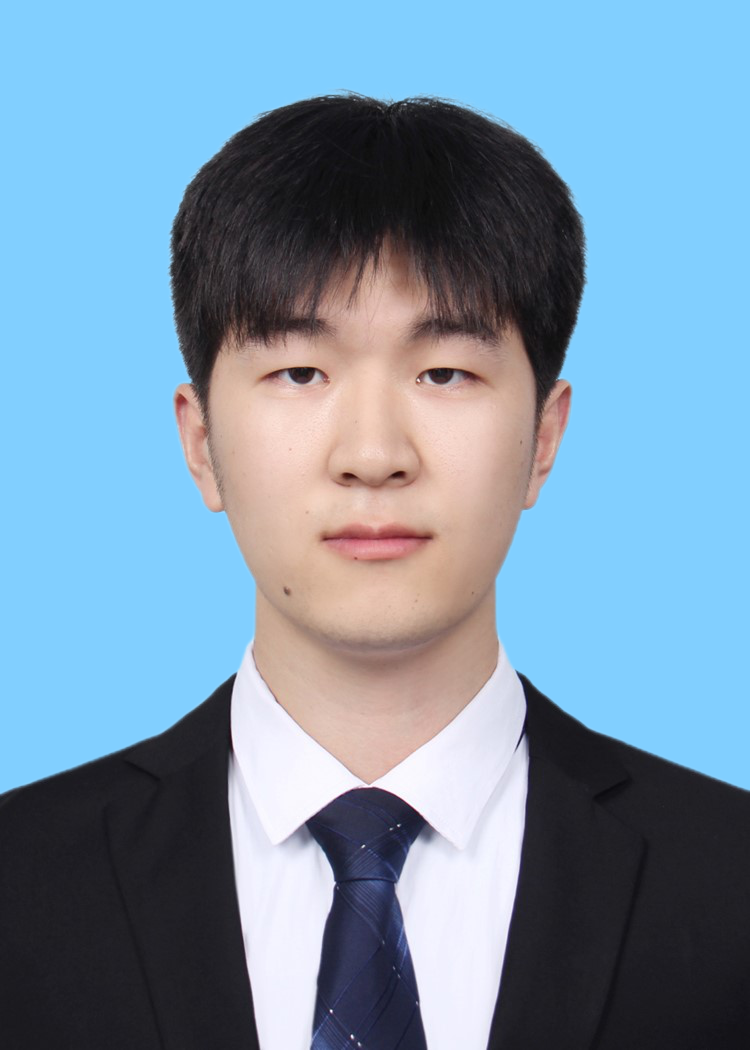}}]{Zhixiang Wang}
	received the M.E. degree in control theory and engineering from Harbin Institute of Technology, Harbin, China, in 2021. He is currently pursuing the Ph.D. degree in Intelligent Science and Technology in the School of Artificial Intelligence and Automation at Huazhong University of Science and Technology, Wuhan, China. His research focuses primarily on model predictive control, motion planning, and reinforcement learning in mobile manipulation legged robots.
\end{IEEEbiography}

\vspace{-1.5cm}
\begin{IEEEbiography}[{\includegraphics[width=1in, height=1.25in, clip, keepaspectratio]{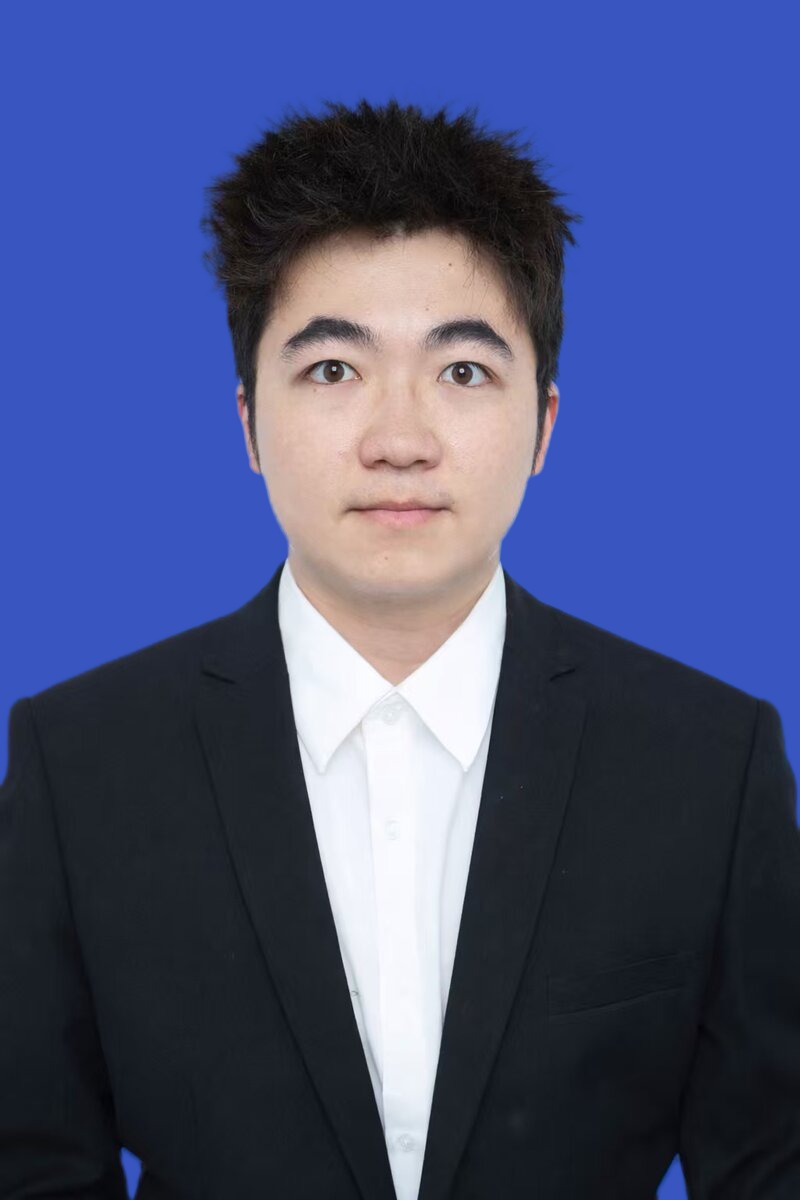}}]{Xiaotian Ke}
	received the B.E. degree in automation from Northeastern University, Shenyang, China, in 2023. He is currently pursuing the M.E. degree in control science and engineering at the School of Artificial Intelligence and Automation, Huazhong University of Science and Technology, Wuhan, China. His research interests focus on robot control.
\end{IEEEbiography}

\vspace{-13.5cm}
\begin{IEEEbiography}[{\includegraphics[width=1in, height=1.25in, clip, keepaspectratio]{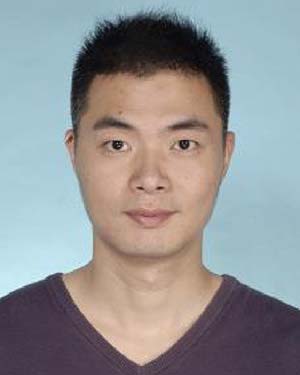}}]{Lijun Zhu}
	received the Ph.D. degree in Electrical Engineering from the University of Newcastle, Callaghan, Australia, in 2013. He is currently a Professor in the School of Artificial Intelligence and Automation at Huazhong University of Science and Technology, Wuhan, China. Prior to this, he was a post-doctoral fellow at the University of Hong Kong and the University of Newcastle. His research interests include power systems, multi-agent systems, and nonlinear systems analysis and control.
\end{IEEEbiography}

\vspace{-13.5cm}
\begin{IEEEbiography}[{\includegraphics[width=1in, height=1.25in, clip, keepaspectratio]{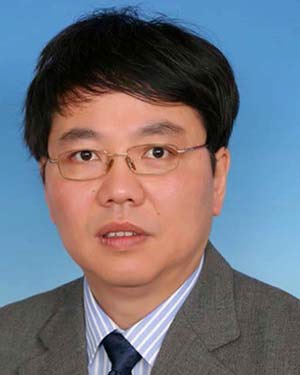}}]{Han Ding}
	received the Ph.D. degree in mechanical engineering from the Huazhong University of Science and Technology (HUST), Wuhan, China, in 1989. 
	
	Supported by the Alexander von Humboldt Foundation, he was a Researcher with the University of Stuttgart, Stuttgart, Germany, from 1993 to 1994. He has been a professor with HUST since 1997. He was a Cheung Kong Chair Professor at Shanghai Jiao Tong University from 2001 to 2006. His research interests include intelligent manufacturing and robotic machining.
	
	Dr. Ding was elected a member of the Chinese Academy of Sciences in 2013.
\end{IEEEbiography}

\end{document}